\documentclass[]{article}
\usepackage{arxiv}

\usepackage{amsfonts}
\usepackage{latexsym}
\usepackage{tabularx}
\usepackage{graphicx}
\usepackage{amssymb}
\usepackage[inline, shortlabels]{enumitem}
\usepackage{amsmath,mathtools}
\newcommand{\qed}{\hfill \ensuremath{\Box}}

\newcommand{\handout}[5]{
   \renewcommand{\thepage}{\arabic{page}}
   \noindent
   \begin{center}
   \framebox{
      \vbox{
        \hbox to 5.78in { {\bf M.Stat Project work} \hfill #2 }
        \vspace{4mm}
        \hbox to 5.78in { { \hfill #5  \hfill} }
        \vspace{4mm}
        \hbox to 5.78in { { #3 \hfill #4}  
        }
      }
   }
   \end{center}
   \vspace*{4mm}
}

\newlength{\rowidth}
\AtBeginDocument{\setlength{\rowidth}{3em}}

\usepackage{tikz}
\usetikzlibrary{positioning}
\newdimen\nodeDist
\usepackage{caption}
\usepackage[linguistics]{forest}

\usepackage{collectbox}

\makeatletter

\usepackage{tcolorbox}

\usepackage{multirow}
\usepackage{tabto}
\usepackage{pgfplots}
\pgfplotsset{width=10cm,compat=1.9}

\usepackage{array}

\usepackage[affil-it]{authblk}
\usepackage[english]{babel}
\usepackage{blindtext}
\usepackage[T1]{fontenc}
\usepackage{amsmath}
\usepackage{amssymb}
\usepackage{multirow}
\usepackage{tabu}
\usepackage{booktabs}
\usepackage[utf8]{inputenc} 
\usepackage{url}            
\usepackage{amsfonts}       
\usepackage{nicefrac}       
\usepackage{microtype}      
\usepackage{lipsum}
\usepackage{floatrow}
\usepackage{graphicx}
\usepackage{hyperref}
\usepackage{wrapfig}
\usepackage[normalem]{ulem}

\usepackage[label font=bf,labelformat=simple]{subfig}
\usepackage{caption}
\title{CMET: Clustering guided METric for quantifying embedding quality}

\author[1]{Sourav Ghosh}
\author[2]{Chayan Maitra}
\author[2,*]{Rajat K. De}

\affil[1]{Standard Chartered Modelling And Analytics Centre Private Limited, Bangalore 560095, India.}
\affil[2]{Machine Intelligence Unit, Indian Statistical Institute, 203 Barrackpore Trunk Road, Kolkata 700108, India.}

\affil[*]{Corresponding author: Rajat K. De, rajat@isical.ac.in}

\begin{document}

\maketitle
\begin{abstract}
Due to rapid advancements in technology, datasets are available from various domains. In order to carry out more relevant and appropriate analysis, it is often necessary to project the dataset into a higher or lower dimensional space based on requirement. Projecting the data in a higher-dimensional space helps in unfolding intricate patterns, enhancing the performance of the underlying models. On the other hand, dimensionality reduction is helpful in denoising data while capturing maximal information, as well as reducing execution time and memory.In this context, it is not always statistically evident whether the transformed embedding retains the local and global structure of the original data. Most of the existing metrics that are used for comparing the local and global shape of the embedding against the original one are highly expensive in terms of time and space complexity. In order to address this issue, the objective of this study is to formulate a novel metric, called Clustering guided METric (CMET), for quantifying embedding quality. It is effective to serve the purpose of quantitative comparison between an embedding and the original data. CMET consists of two scores, viz., $CMET_L$ and $CMET_G$, that measure the degree of local and global shape preservation capability, respectively. The efficacy of CMET has been demonstrated on a wide variety of datasets, including four synthetic, two biological, and two image datasets. Results reflect the favorable performance of CMET against the state-of-the-art methods. Capability to handle both small and large data, low algorithmic complexity, better and stable performance across all kinds of data, and different choices of hyper-parameters feature CMET as a reliable metric.  
\end{abstract}




\section{Introduction}
\label{sec:intro}

Transformation of the data leading to change in dimensionality poses several challenges. Especially handling high-dimensional data is a longstanding issue in data science \cite{highd_probs}. As the dimension increases, the dataset becomes increasingly sparse, making it harder to identify hidden patterns—a phenomenon known as the "curse of dimensionality" \cite{curseofDim}. High dimensional data necessitates a large number of training samples to effectively train models on high-dimensional data. A widely adopted solution is Dimensionality Reduction (DR), which aims to project the data into a lower-dimensional space while minimizing information loss relative to the original dataset \cite{dimRed}. Existing literature shows a wide variety of dimension reduction techniques based on different notions, e.g., linear, non-linear, and manifold-based approaches. The models differ methodologically, and each method is devised to serve diverse purposes. Linear methods, like PCA (Principal Component Analysis), SVD (Singular Value Decomposition), and Factor Analysis \cite{inear_DR_Methods}, are predominantly popular for their information retention capabilities, whereas non-linear methods, like TSNE (t-Distributed Stochastic Neighbor Embedding), Kernel PCA, UMAP (Uniform Manifold Approximation and Projection) \cite{non-lin_DR_review} and IVIS \cite{IVIS}, are concerned more about preserving intricate patterns of the original data in lower dimensional embedding. It may be mentioned here that UMAP is a manifold-based technique; IVIS is a neural network-based technique, TSNE and FITSNE (Fast Interpolation-based t-Distributed Stochastic Neighbor Embedding) are distribution-based techniques, which are designed to preserve local and global structures in high-dimensional data while enabling effective visualization and clustering \cite{review_DR} \cite{dimred_review_2}. Instead of having a rigorous mathematical foundation, these methods, to a great extent, justifies their accomplishment for serving certain purposes, without any sensible quantification. It is not very clear to the users to rely upon any particular method to use for reducing the dimensionality of the data in hand. 

Since it has been a longstanding challenge in related domains, numerous measures have been developed to quantify the quality of embeddings. Some of the most widely used approaches are discussed here. One common approach, Distance Matrix, focuses on preserving pairwise distances among all data points \cite{yang2011distance}. It involves constructing two matrices: one for the original high-dimensional data and another for the lower-dimensional embedding. The quality of the embedding is then assessed by computing the Spearman’s or Pearson’s correlation between these matrices. Another commonly used approach involves rank-based methods, which utilize the distance matrix to define new evaluation measures. A ranking matrix is derived from the distance matrix, where each element represents the neighborhood-wise rank of a data point. The co-ranking matrix \cite{co-ranking/co-k} is an extension of the ranking matrix, with its $(i,j)^{th}$ element indicating number of data points of rank $i$ in the original space attaining rank $j$ in the embedded space.

Trustworthiness and continuity \cite{trust} are two measures based on a co-ranking matrix. If some far-away data points are pulled close by a DR method, it is called "intrusion". Similarly, if some close data point is pushed away, it is called "extrusion". Both these measures are based on certain number of hard extrusions, which is nothing but function of elements of co-ranking matrix, properly scaled and adjusted by scaler so that both the measures take values in $[0,1]$, where higher value indicates better shape preservation. Co-k-nearest neighbor size and local continuity meta criterion (LCMC) \cite{LCMC} are also similar measures derived from the co-ranking matrix.

Manifold embedding quality assessment \cite{MEQA} is another metric that incorporates certain transformations, like translation, anisotropic scaling, and rotation, to evaluate the degree of distortion of the arrangement of the neighbourhood structure made by the DR method under consideration. To infer about global shape preservation, it also uses the Kolmogorov-Smirnov test. On the contrary, Normalization Independent Embedding Quality Assessment (NIEQA) \cite{NIEQA} provides both local and global metric based on anisotropic scaling independent measure. In a review paper by Zhang et al \cite{pydrm}, some of the relevent rank basesd metrics are reviewed and the authors also provide a package in python, called "pyDRMetrics" which offers an easy computational pathway to those metrics.

Most of the measures are designed mainly to evaluate the local shape-preserving power of the DR methods since global shape preservation is less important in most of the cases. Also, being unable to handle large sized data and taking significant amount of run-time are two remarkable drawbacks of those measures. A dataset comprising $n$ samples will produce an $n\times n$ distance matrix, and consequently the corresponding co-ranking matrix will be of the same order. Naturally, all the metrics derived from the co-ranking matrix have to work with $n^{2}$ elements of the co-ranking matrix. Thus, we have needfully designed a metric, which reflects the capability of any transformation, e.g., a DR method, in both local and global shape preservation.

In this study, we introduce a metric CMET (Clustering guided METric) that can quantify the quality of any embedding produced by any transformation in terms of their efficacy in preserving the local and global structure of the data. The proposed metric, CMET, uses unsupervised clustering technique to characterize the shape of the data set in the original higher dimensional space in terms of  a number of clusters, which provides information about the locality of a sample and also about its location with respect to the scattered-ness of the whole data in that space. An ideal transformation should keep samples in the same globular cluster structure in another dimension so that the points of vicinity in the original space, would stay in vicinity in the transformed space, and vice versa. CMET acknowledges the local and global shape preservation by a transformation in two parallel ways. CMET comprises two numeric scores, viz., $CMET_L$ and $CMET_G$ respectively, which quantify local and global shape preserving capability of any transformation. Furthermore, we describe how it overcomes the drawbacks of the prevalent metrics, and successfully addresses all the requirements for understanding the nature of the lower dimensional embedding. To achieve reasonable evidence on the accountability of CMET, a total of $14$ state-of-the-art DR methods and $8$ datasets of different kinds have been used demonstrating performance of CMET standalone, and also as a part of a performance comparison study with other metrics.

The outline of the present article is as follows. Section \ref{sec:methodology} describes the motivation behind the proposed metric (CMET) with a detailed description of the mathematical approach. In addition, the supervised and unsupervised concept of local and global measures used in CMET is discussed elaborately. In Section \ref{sec:Results}, all related results are discussed, which sheds light on different aspects of CMET. This section includes the efficacy of CMET, the sensitivity analysis of the hyper-parameter, and a comparison of CMET's efficiency against the state-of-the-art metrics. Section \ref{sec:DisCon} describes the result discussion, and a conclusion which includes the limitation and future scopes.


\section{Methodology}
\label{sec:methodology}
For the sake of analyzing the data in a more appropriate way, it is often necessary to transform the whole data set in some other dimension. In particular, high-dimensional datasets often contain noise and irrelevant information, and it is also difficult to build simple interpretable models on data sets with a large number of features. A sophisticated statistical approach to address the issue is to reduce the dimensionality of the data to get an embedding, which is reasonably free from unintentional noise and more concisely present only the relevant information. However, the availability of multiple competent DR methods creates confusion for the user. The lack of a potentially reliable metric capable of quantifying how good a DR method is at preserving the local and global shape of the original data is the central motivation of this study. 


Thus, in this study, a clustering-guided metric, called CMET, has been developed. CMET is far less computationally expensive than other state-of-the-art metrics in terms of time and space complexity. At the preliminary stage, in the formulation of CMET, the dataset has been divided into multiple clusters so that a notion about the distribution and spread of sample points in the same dimensional space is obtained. Clustering also helps in understanding the inherent local relationships present in the dataset. Agglomerative clustering has been considered to serve this purpose. 

The central idea of the formulation of CMET lies behind the fact that the points that are forming clusters in the original space, will also stay close to each other in the transformed embedding, and the clusters will maintain similar relative distance among them, preserving the local and global shape, respectively, of the original data in the transformed embedding.

However, for labeled data, one may use the data labels instead of clustering the data by some unsupervised clustering method, as the classes often constitute well separable clusters in the data set.  Use of present class labels as clusters makes the approach supervised and consequently reduces the computational complexity of CMET bypassing Agglomerative clustering, which has a complexity of the order $\mathcal{O}(n^2logn)$.  However, to ensure unambiguity and consistency, the methodology described below is aligned with unsupervised approach of CMET. In the result section, CMET scores are presented for both supervised and unsupervised methods.


CMET comprises two scores, $CMET_L$ and $CMET_G$. $CMET_L$, signifying extent of local shape preservation, is  defined based on the pairwise distances of the samples from centroids of the clusters which the samples belong to, both in the original and transformed spaces. On the other hand, the measure for global shape preservation, $CMET_G$, is formulated based on the pairwise distances among the centroids of the clusters in the original and transformed spaces.

Consider a dataset in the original space, comprising $n$  $p$-dimensional samples distributed in $c$ clusters. After transformation, the dimensionality of the data is changed from $p$ to $q$. We form the same set of clusters, after transformation, by putting the samples in the respective clusters in the transformed space. Henceforth, the following notations are considered throughout the study. Let $\mathbf{X}_{n \times p} = \{\mathbf{x}_i\}_{i=1}^n$ be the original dataset, and $\mathbf{X}^{'}_{n \times q} = \{\mathbf{x}'_i\}_{i=1}^n$ be the corresponding transformed embedding. Let $\mathbf{y}$ be an n-dimensional vector, such that its $i^{th}$ component $\mathbf{y}_i$ is either the class label of $i^{th}$ sample or its cluster label obtained by a clustering algorithm. For notational convenience, we would like to introduce the following equivalence : $\{\mathbf{x}_i\}_{i=1}^{n} \equiv  \{\mathbf{z}_{j(k)}\}$; $\{\mathbf{x}_i^{'}\}_{i=1}^{n} \equiv  \{\mathbf{z}_{j(k)}^{'}\}$, where $j(k)=1,2,\cdots,m(k)$; $k=1,2,\cdots,c$; $n=\sum_{k=1}^c m(k)$, such that $\mathbf{z}_{j(k)}$ and $\mathbf{z}_{j(k)}^{'}$ be the $j^{th}$ sample belonging to $k^{th}$ cluster in the original and transformed spaces respectively. Let, $\boldsymbol\mu_{k}$ and $\boldsymbol{\mu}_{k}^{'}$ denote the median of the samples belonging to the $k^{th}$ cluster in the original and transformed spaces respectively. Also let us define $S_{k}$, the set of all the samples present in the $k^{th}$ cluster in the original space, i.e., $S_{k} = \{\mathbf{z}_1, \mathbf{z}_2,\cdots, \mathbf{z}_{m(k)}\}$. Likewise, we define $S_{k}^{'} = \{\mathbf{z}_1^{'}, \mathbf{z}_2^{'},\cdots, \mathbf{z}_{m(k)}^{'}\}$ in the transformed space.

\subsection{\texorpdfstring{$CMET_L$}{CMET\_L}}
Let us define $\mathbf{r}=  \begin{bmatrix} r_{k} \end{bmatrix}_{k=1}^{c}$ for the original dataset, such that
\begin{equation*}
r_k = 
\begin{cases}
\underset{j(k)}{\max} || \mathbf{z}_{j(k)}-\boldsymbol{\mu}_{k}|| & \text { if } {\exists  \ j(k) :||\mathbf{z}_{j(k)}-\boldsymbol{\mu}_{k}|| > 0,  \text{ where } j(k)=1,2,\cdots,m(k) } \\
1 & \text{otherwise,}
\end{cases}
\end{equation*}
where $||\cdot||$ is the Euclidean norm.
Similarly, we define $\mathbf{r^{'}}=\begin{bmatrix}
 r_{k}^{'}  
  \end{bmatrix}_{k=1}^{c}$, such that
\begin{equation*}
r_k^{'} = \begin{cases}
\underset{j(k)}{\max} || \mathbf{z}_{j(k)}^{'}-\boldsymbol{\mu}_{k}^{'}|| & \text { if } {\exists  \ j(k) :||\mathbf{z}_{j(k)}^{'}-\boldsymbol{\mu}_{k}^{'}|| > 0,  \text{ where } j(k)=1,2,\cdots,m(k) } \\
1 & \text{otherwise}
\end{cases}
\end{equation*}
The terms $\mathbf{r}$ and $\mathbf{r}^{'}$ signify the extent of $c$ clusters in the original and transformed spaces. Now two vectors $\mathbf{d}=  \begin{bmatrix} d_{i} \end{bmatrix}_{i=1}^{n}$ and $\mathbf{d}^{'}=  \begin{bmatrix} d_{i}^{'} \end{bmatrix}_{i=1}^{n}$, for the original data and the transformed embedding respectively,  are defined as
$$d_i= \frac{|| \mathbf{z}_{j(k)}-\boldsymbol{\mu}_{k}||}{r_{k}}; \forall i =1,2,\cdots,n; \text{ where } \{x_i\}_{i=1}^{n} \equiv  \{\mathbf{z}_{j(k)}\}, j(k)=1,2,\cdots,m(k)\ \text{and}\  k=1,2,\cdots,c .$$
$$d_i^{'} = \frac{|| \mathbf{z}_{j(k)}^{'}-\boldsymbol{\mu}_{k}^{'}||}{r_{k}^{'}};  \forall i =1,2,\cdots,n ; \text{ where} \ \{x_i^{'}\}_{i=1}^{n} \equiv  \{\mathbf{z}_{j(k)}^{'}\}, j(k)=1,2,\cdots,m(k)\ \text{and}\  k=1,2,\cdots,c .$$ 
The term $\mathbf{d}$ and $\mathbf{d}^{'}$ reflects the normalized distances of the samples from the centroid of the respective clusters which the samples belong to, both in the original and transformed spaces. With the help of the above, local CMET score $CMET_L$ is defined as

$$CMET_L = 1- \frac{|| \mathbf{d}-\mathbf{d}^{'}||}{\sqrt{n}}$$
 
The value of $CMET_L$ lies in $[0,1]$, the proof of which is given below. The rationale behind the construction of the proposed metric is the following. If a DR method can preserve the local and global shape of the data, then even after dimensionality reduction, closer samples remain closer, distant samples remain distant in the embedding as well. In terms of local shape preservation, ideally both distance vectors $\mathbf{d}$ and $\mathbf{d}^{'}$ should contain entries which are numerically closer. On the other hand, if the DR method fails to preserve the shape of the data, the entries corresponding to the ${i}^{th}$ sample in $\mathbf{d}$ and $\mathbf{d}^{'}$ may be quite different. 
Therefore, the Euclidean distance between $\mathbf{d}$ and $\mathbf{d}^{'}$ would be close to $0$ if the embedding is good, and large otherwise. For $n$ samples, $ || \mathbf{d}-\mathbf{d}^{'}||$ is scaled so that $ \frac{|| \mathbf{d}-\mathbf{d}^{'}||}{\sqrt{n}}$ does not exceed $1$. Thus, higher value of $CMET_L$ signifies better preservation of local shape and vice versa.


\textbf{Corollary $1$:} The local score, $CMET_L$, always lies in $[0, 1]$.

\textbf{Proof}: With the above notation, we have


$0 \leq || \mathbf{z}_{j(k)}-\boldsymbol{\mu}_{k}|| \leq {r_k}$ (by definition of $r_k$)  

$\implies 0 \leq \frac{|| \mathbf{z}_{j(k)}-\boldsymbol{\mu}_{k}||}{r_k} \leq 1 $  

$\implies 0 \leq d_{i} \leq 1 $

Analogously, we get, $0 \leq \frac{|| \mathbf{z}_{j(k)}^{'}-\boldsymbol{\mu}_{k}^{'}||}{r_k^{'}} = d_{i}^{'} \leq 1 $ 

Now, for the vector, 
$\mathbf{d} - \mathbf{d}^{'}$=$\begin{bmatrix}
d_{i} - d_{i}^{'}  \end{bmatrix}_{i=1}^{n}$, 

note that, $-1 \leq d_{i} - d_{i}^{'} \leq 1$ $ \forall i = 1,2,\cdots,n$

$\implies 0 \leq (d_{i} - d_{i}^{'})^2 \leq 1 \implies 0 \leq 
 \sqrt{\sum_{i=1}^n (d_{i} - d_{i}^{'})^2 } \leq \sqrt{n} \implies 0 \leq 
 || \mathbf{d} - \mathbf{d}^{'} || \leq \sqrt{n}$  

Therefore, $0 \leq CMET_L = 1 - \frac{|| \mathbf{d} - \mathbf{d}^{'} ||}{\sqrt{n}} \leq 1$


\subsection{\texorpdfstring{$CMET_G$}{CMET\_G}}
Let $\boldsymbol{\mu}$ and $\boldsymbol{\mu}^{'}$ be the medians of the entire original data and the transformed data respectively. Without loss of generality and continuation of the previous notations, let $\boldsymbol{\mu}_{c+1} = \boldsymbol{\mu}$ and $\boldsymbol{\mu}_{c+1}^{'} = \boldsymbol{\mu}^{'}$. 

Let us consider two median distance matrices $[|| \boldsymbol{\mu}_{k} - \boldsymbol{\mu}_{l}||]_{(c+1) \times (c+1)}$, and $[|| \boldsymbol{\mu}_{k}^{'} - \boldsymbol{\mu}_{l}^{'}||]_{(c+1) \times (c+1)}$ in the original and transformed spaces respectively.

We define
\begin{equation*}
\mathbf{\Gamma}=\{\gamma_{kl}\}_{(c+1)\times (c+1)}=\begin{cases}
\frac{\{|| \boldsymbol{\mu}_{k} - \boldsymbol{\mu}_{l}||\}_{(c+1) \times (c+1)}}{\max_{k,l}|| \boldsymbol{\mu}_{k} - \boldsymbol{\mu}_{l}||} & \text{if }{\max_{k,l}|| \boldsymbol{\mu}_{k} - \boldsymbol{\mu}_{l}|| \neq 0} \\
\mathbf{O}_{(c+1) \times (c+1)} & \text {otherwise}, 
\end{cases}
\end{equation*} 
and 
\begin{equation*}
\mathbf{\Gamma}^{'}=\{\gamma^{'}_{kl}\}_{(c+1)\times (c+1)}=\begin{cases}
\frac{\{|| \boldsymbol{\mu}_{k}^{'} - \boldsymbol{\mu}_{l}^{'}||\}_{(c+1) \times (c+1)}}{\max_{k,l}|| \boldsymbol{\mu}_{k}^{'} - \boldsymbol{\mu}_{l}^{'}||} & \text{if }{\max_{k,l}|| \boldsymbol{\mu}_{k}^{'} - \boldsymbol{\mu}_{l}^{'}|| \neq 0} \\
\mathbf{O}_{(c+1) \times (c+1)} & \text {otherwise} 
\end{cases}
\end{equation*}

The elements of the matrices $\mathbf{\Gamma}$ and $\mathbf{\Gamma}^{'}$, corresponding to the original and transformed space, reflect the relative pairwise distances between the centroid medians of the clusters including that of the entire data.

We introduce global measure of CMET as 

$$CMET_G = 1-\frac{||\mathbf{\Gamma}-\mathbf{\Gamma}^{'}||_F}{\sqrt{c(c+1)}},$$ 
where $||\cdot||_F$ is the Frobinius norm of a matrix. Like $CMET_L$, $CMET_G$ also lies in $[0,1]$. The proof of which is given below. Here, global shape is described as the inter-cluster relationship. All the cluster medians along with the global median are taken to be the representatives of the clusters and the whole dataset, respectively. For either of the original or transformed spaces, all the pair-wise distances are stored in a matrix, which is scaled by the maximum element of the median distance matrix to have all the elements of the distance matrix lying in $[0,1]$. Just like the local shape measure, there are also two distance matrices, $\mathbf{\Gamma}$ and $\mathbf{\Gamma}^{'}$, for the original and the transformed embedding. Taking the Frobinius norm of the difference between the two distance matrices, a quantification is achieved regarding how much the medians of the clusters are deviated from the original structure of the data. Thus, higher the value of $CMET_G$, better is the preservation of global shape, and vice versa. 


\textbf{Corollary $2$}: The global score, $CMET_G$, always lies in $[0,1]$.

\textbf{Proof} : Total number of elements in a distance matrix = $(c+1)^2$. 

Number of diagonal elements = $c+1$ 

So, maximum number of non-zero elements = $(c+1)^2-(c+1)$$=c(c+1)$ 

Now, note that, 

$|| \boldsymbol{\mu}_{k} - \boldsymbol{\mu}_{l}|| \geq 0$ $\forall k,l = 1,2,\cdots,c+1.$ 
$\implies 0\leq \frac{|| \boldsymbol{\mu}_{k} - \boldsymbol{\mu}_{l}||}{\max_{k,l}|| \boldsymbol{\mu}_{k} - \boldsymbol{\mu}_{l}||}\leq 1 $ 

Similarly, we get, 
$ 0\leq \frac{|| \boldsymbol{\mu}_{k}^{'} - \boldsymbol{\mu}_{l}^{'}||}{\max_{k,l}|| \boldsymbol{\mu}_{k}^{'} - \boldsymbol{\mu}_{l}^{'}||}\leq 1 $ 

From the above two inequalities we have $0 \leq \gamma_{kl},\gamma_{kl}^{'} \leq 1 \implies -1 \leq \gamma_{kl} - \gamma_{kl}^{'}\leq 1 \implies 0 \leq (\gamma_{kl}-\gamma_{kl}^{'})^2\leq 1$ 

Now, $||\mathbf{\Gamma}-\mathbf{\Gamma}^{'}||_F=||\{\gamma_{kl}-\gamma_{kl}^{'}\}_{(c+1) \times (c+1)}||_F$ $=\sqrt{{\sum_{k=1}^{c+1}\sum_{l=1}^{c+1} (\gamma_{kl} - \gamma_{kl}^{'})^2 }} \leq \sqrt{c(c+1)}$ 

Therefore, $0 \leq CMET_G = 1-\frac{||\mathbf{\Gamma}-\mathbf{\Gamma}^{'}||_F}{\sqrt{c(c+1)}}\leq 1$ 



\section{Results}
\label{sec:Results}

CMET is devised to quantitatively figure out the best DR method which is able to reproduce the original local and global shape in the embedding produced by it. A total of fourteen methods have been used for comparison and are as follows: 
\begin{enumerate*}[series = tobecont, itemjoin = \quad]
\item PCA \cite{PCA}, \item SVD \cite{SVD}, \item ICA \cite{ICA}, \item ISOMAP \cite{ISOMAP}, \item FA \cite{FA}, \item NMF \cite{NMF}, \item LLE \cite{LLE}, \item LDA \cite{LDA}, \item TSNE \cite{TSNE}, \item FITSNE \cite{FITSNE}, \item KPCA \cite{KPCA}, \item UMAP \cite{UMAP}, \item PHATE \cite{PHATE}, and \item IVIS \cite{IVIS}.
\end{enumerate*}
To justify CMET's robustness and its wide as well as unbiased applicability emperically, several kinds of datasets have been used. A brief description of the datasets is tabulated in Table \ref{tab:my_label} :

\begin{table}[]
    \centering
    \begin{tabular}{p{1.7cm}p{5cm}p{1.5cm}p{1.4cm}p{1.4cm}p{1.2cm}p{1.1cm}}
    \hline
    \begin{center}\textbf{Dataset} \end{center} & \begin{center}\textbf{Description} \end{center}& \begin{center}\textbf{Type} \end{center}& \begin{center}\textbf{Samples} \end{center}& \begin{center}\textbf{Features} \end{center}& \begin{center}\textbf{Classes} \end{center}& \begin{center}\textbf{Source} \end{center}\\ \hline 
    \begin{center}$Olympics$ \end{center}& \begin{center}Five interlaced ring shaped data \end{center}& \begin{center}Synthetic \end{center} &  \begin{center}$2500$ \end{center}& \begin{center}$2$ \end{center}& \begin{center}$5$ \end{center}& \begin{center}\cite{ndavis}   \end{center}\\[-2ex]
    \begin{center}$WorldMap$ \end{center}& \begin{center}Five large continent shaped data \end{center}& \begin{center} Synthetic \end{center} & \begin{center}$2843$ \end{center}& \begin{center}$2$ \end{center}& \begin{center}$5$ \end{center}& \begin{center}\cite{ndavis}   \end{center}\\[-2ex]
    \begin{center}$Shape$ \end{center}& \begin{center} Data in the form of the letters S,H,A,P, and E \end{center} & \begin{center} Synthetic \end{center} & \begin{center}$2000$ \end{center}& \begin{center}$2$ \end{center}& \begin{center}$5$ \end{center}& \begin{center}\cite{ndavis}   \end{center}\\[-2ex]
    \begin{center}$Swiss Roll$ \end{center}& \begin{center}It consists of points arranged in a "roll" shape \end{center} & \begin{center} Synthetic \end{center} & \begin{center}$1500$ \end{center}& \begin{center}$3$ \end{center} & \begin{center}$4$ \end{center}& \begin{center}\cite{SwissRoll}  \end{center}\\ [-2ex]\hline
    \begin{center}$Jurkat$ \end{center}& \begin{center} RNA-seq obtained from the peripheral blood of a 14-year-old boy with acute lymphoblastic leukemia \end{center} & \begin{center}Biological \end{center} & \begin{center}$3364$ \end{center}& \begin{center}$2423$ \end{center}& \begin{center}$12$ \end{center}& \begin{center}\cite{jurkat}   \end{center}\\[-2ex]
    \begin{center}$Zeisel$ \end{center}& \begin{center}RNA-seq dataset collected from a mouse brain single cells \end{center} & \begin{center} Biological \end{center} & \begin{center}$3004$ \end{center}& \begin{center}$4041$ \end{center} & \begin{center}$25$ \end{center} & \begin{center} \cite{zeisel} \end{center} \\ [-2ex]\hline
    \begin{center}$Mnist$ \end{center} & \begin{center}grayscale images of handwritten digits \end{center} & \begin{center}Image \end{center} & \begin{center}$60000$ \end{center}& \begin{center}$784$ \end{center}& \begin{center}$10$ \end{center}& \begin{center}\cite{MNIST}   \end{center}\\[-2ex]
    \begin{center}$FMnist$ \end{center}& \begin{center}grayscale images of clothing items \end{center}& \begin{center}Image \end{center} & \begin{center}$60000$ \end{center}& \begin{center}$784$ \end{center}& \begin{center}$10$ \end{center}& \begin{center}\cite{FMNIST}   \end{center}\\[-2ex] \hline
    \end{tabular}
    \caption{Brief description of datasets used in this study.}
    \label{tab:my_label}
\end{table}

In this section, we will demonstrate the efficacy of CMET by showing its performance on several datasets (Section \ref{subsec: Efficacy of CMET}), the sensitivity of the hyper-parameter (Section \ref{subsec:Sensitivity of the Hyper-parameter}), and finally a comparison of CMET's efficiency with other state-of-the-art metrics (Section \ref{subsec:Comparison with Other Metrics}).

In Section \ref{subsec: Efficacy of CMET}, dataset-wise results has been described. For each dataset, all the embeddings, produced by dimensionality reduction methods, has been visualized along with the corresponding numerical CMET scores. It is followed by a discussion on how the CMET scores judge the embedding quality correctly. In Section \ref{subsec:Sensitivity of the Hyper-parameter}, the obvious issue with the hyper-parameter of Agglomerative Clustering (which is used to cluster the dataset), viz., number of clusters, has been addressed. The performance of  CMET and the other three metrics, viz., trustworthiness, continuity, and LCMC  has been discussed in the Section \ref{subsec:Comparison with Other Metrics} and found that CMET is much more useful than other state-of the-art metrics in every relevant context.

\subsection{Efficacy of CMET}
\label{subsec: Efficacy of CMET}
Here we describe all the pertinent results related to the application of CMET on different DR methods using the above-mentioned datasets. Datasets from different domains have its own structure and nature. The aim of a DR method is to unfold it, and preserve all those intricate patterns of the original data in the lower dimensional (transformed) embedding. 


\subsubsection{Synthetic datasets}
Use of two or three dimensional synthetic datasets is an elegant way to establish the desired conclusion at the rudimentary level. It paves the way for visual comparison of the embeddings with the original data. Three two-dimensional datasets and one three-dimensional dataset have been considered.

\begin{flushleft}
    \textit{Olympics}
\end{flushleft}

\begin{figure}
\includegraphics[width=\columnwidth]{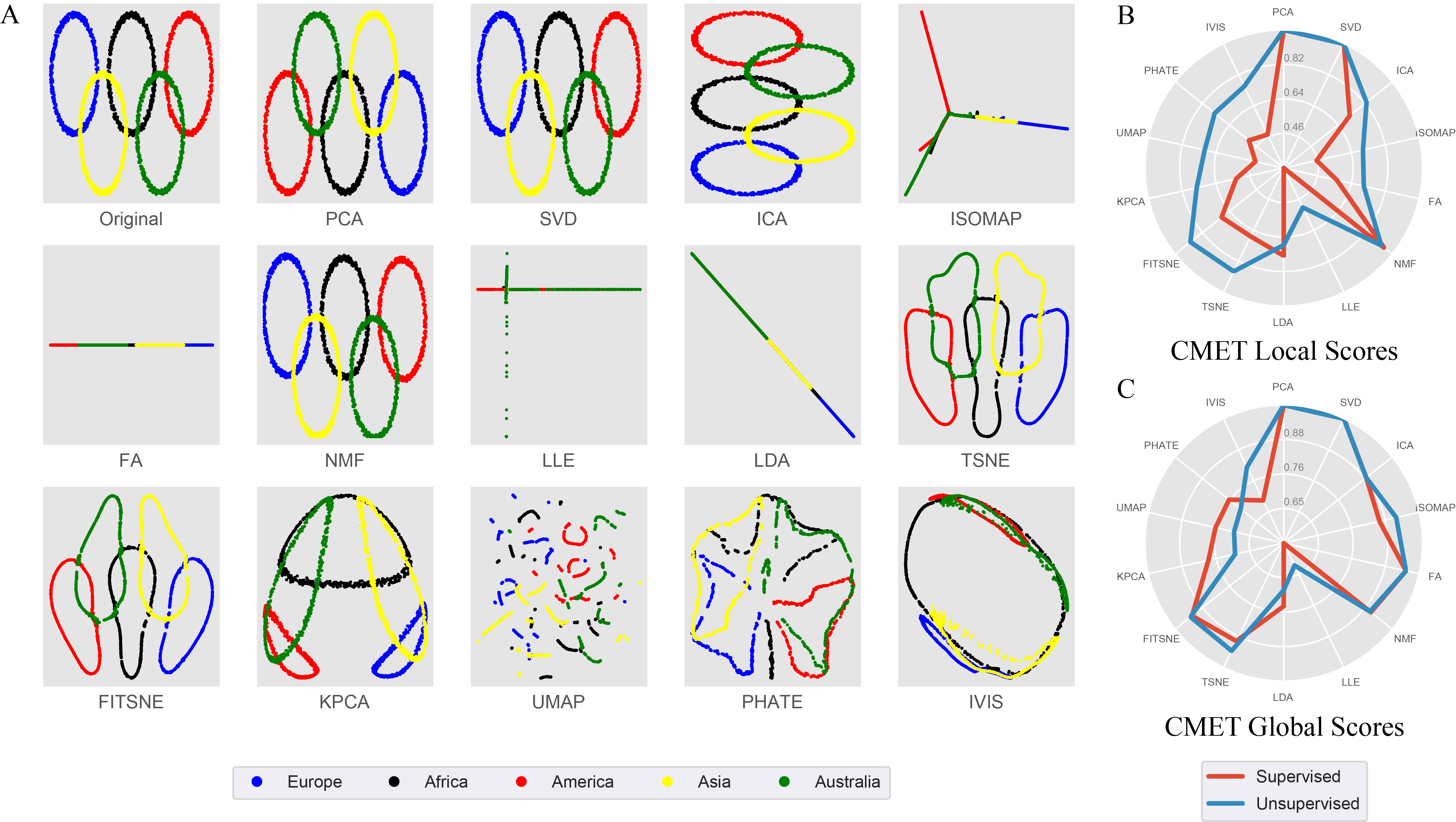}
\caption{Olympics Embeddings and CMET Scores}
\label{fig:Olympics Embeddings and CMET Scores}
\end{figure}

The scatter plot (Figure \ref{fig:Olympics Embeddings and CMET Scores}) of $Olympics$ dataset looks like the Olympic ring, i.e., there are five interlaced rings that symbolize the Olympic movement. We will first visually inspect the embeddings produced by different DR methods, and then analyze the evaluated local and global CMET scores for the embeddings. 

Visual inspections immediately lead in the following observations. NMF and SVD have reproduced almost the original structure. ICA has rotated anti-clock wise and PCA has inverted the original structure. FITSNE and TSNE  have inverted the structure deforming the circular shape of the rings. Other methods have deformed the local and global shapes so drastically that hardly any resemblance with the original structure is found.


On the other hand, CMET scores suggest the following conclusions. CMET score for local shape preservation is very high for PCA, SVD, NMF, which indicates that these DR methods have captured the original local shape of the dataset very well in their resulting embeddings. FITSNE, TSNE and ICA have produced similar embeddings but have not captured the original local shape properly. Their Local CMET score, being lower than the above mentioned methods but higher than many others, also suggest the same. However, $CMET_L$ score for PCA is lower than that of TSNE and FITSNE. It may be due to the fact that interlace rings, though de-shaped, have been better preserved by TSNE and FITSNE than ICA as the orientation of the ring seems to be visually more similar for the first two than ICA. $CMET_L$ scores for FA, UMAP, PHATE, ISOMAP, IVIS, LLE are very poor implying that they could not preserve local shape so well. $CMET_G$ scores of PCA, SVD, NMF, ICA, FA, TSNE, FITSNE and ISOMAP are high since in their embeddings the clusters are located either in the similar fashion like the original or some different order maintaining proper distance from each other irrespective of the shape of the cluster. Other methods are not very significant in Global shape preservation, as suggested by Global CMET score. Hence, the CMET scores have coherence with the visual inspection to a great extent.

\begin{flushleft}
    \textit{WorldMap}
\end{flushleft}

\begin{figure}
\includegraphics[width=\columnwidth]{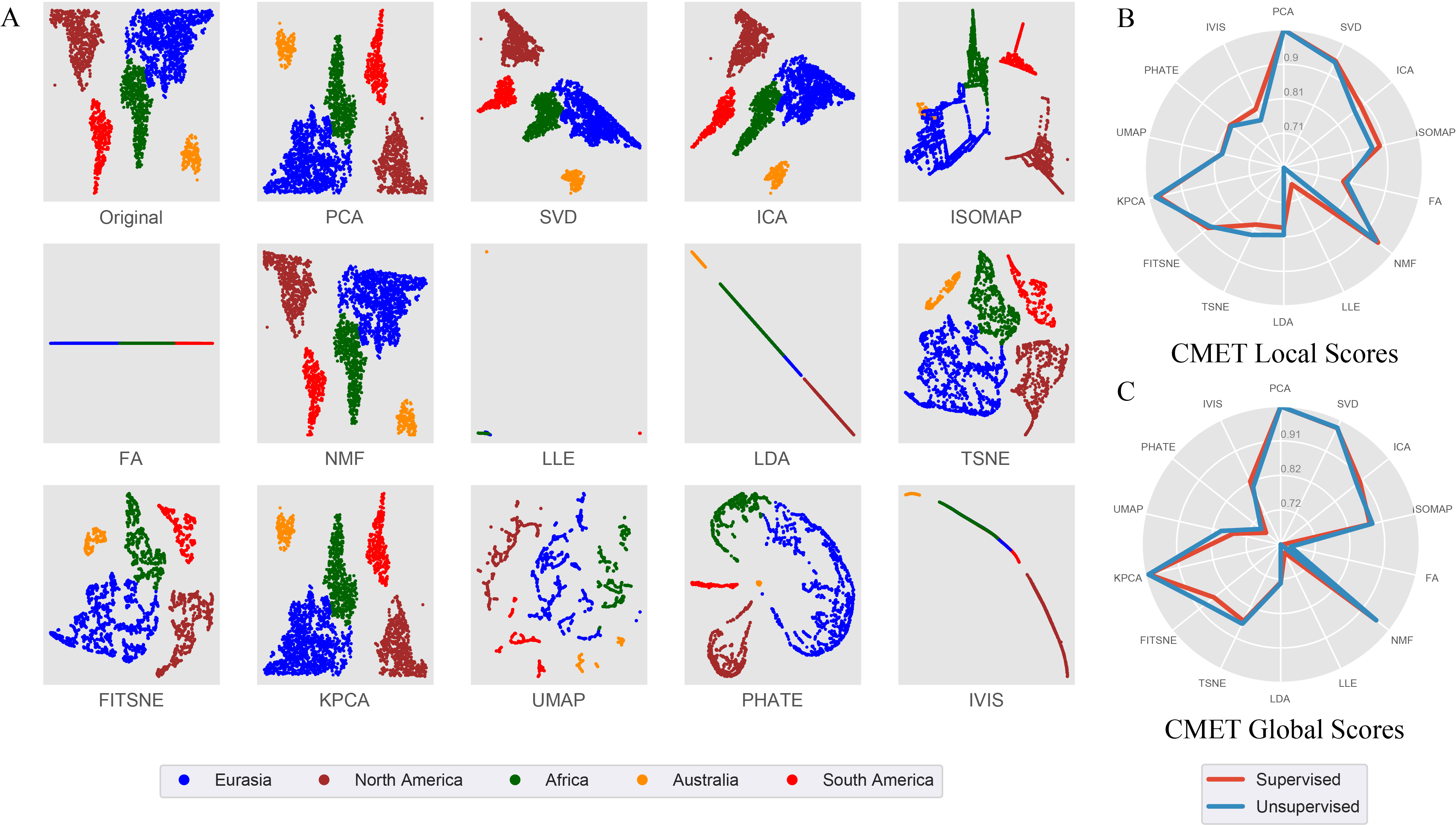}
\caption{WorldMap Embeddings and CMET Scores}
\label{fig:WorldMap Embeddings and CMET Scores}
\end{figure}

In this dataset of the world map (Figure \ref{fig:WorldMap Embeddings and CMET Scores}),  each of five labels corresponds to one of the continents. At first, looking at the embeddings, we will try to detect visually evident deformations of the original structure in the embeddings. The followings are observed. NMF has successfully reproduced almost the original structure of the data. PCA and KPCA have inverted the original structure. FITSNE and TSNE have inverted the original structure but have deformed the local shape. ICA and SVD have rotated the structure clock-wise making slight change in local shape. All other DR methods have deformed the local as well as the global shape to a great extent such that the original structure is not even recognizable in those embeddings.

From the CMET scores for this data, we can get the following observations. PCA, SVD, ICA, NMF, and ISOMAP are performing well in preserving local as well as global shape. FA, LLE, PHATE, and UMAP are poor in preserving both local and global shapes as suggested by the CMET scores since these DR methods could not recover the original data structure. Clearly, the minimally altered embeddings have a greater CMET scores indicating good shape preservation.

\begin{flushleft}
    \textit{Shape}
\end{flushleft}

\begin{figure}
\includegraphics[width=\columnwidth]{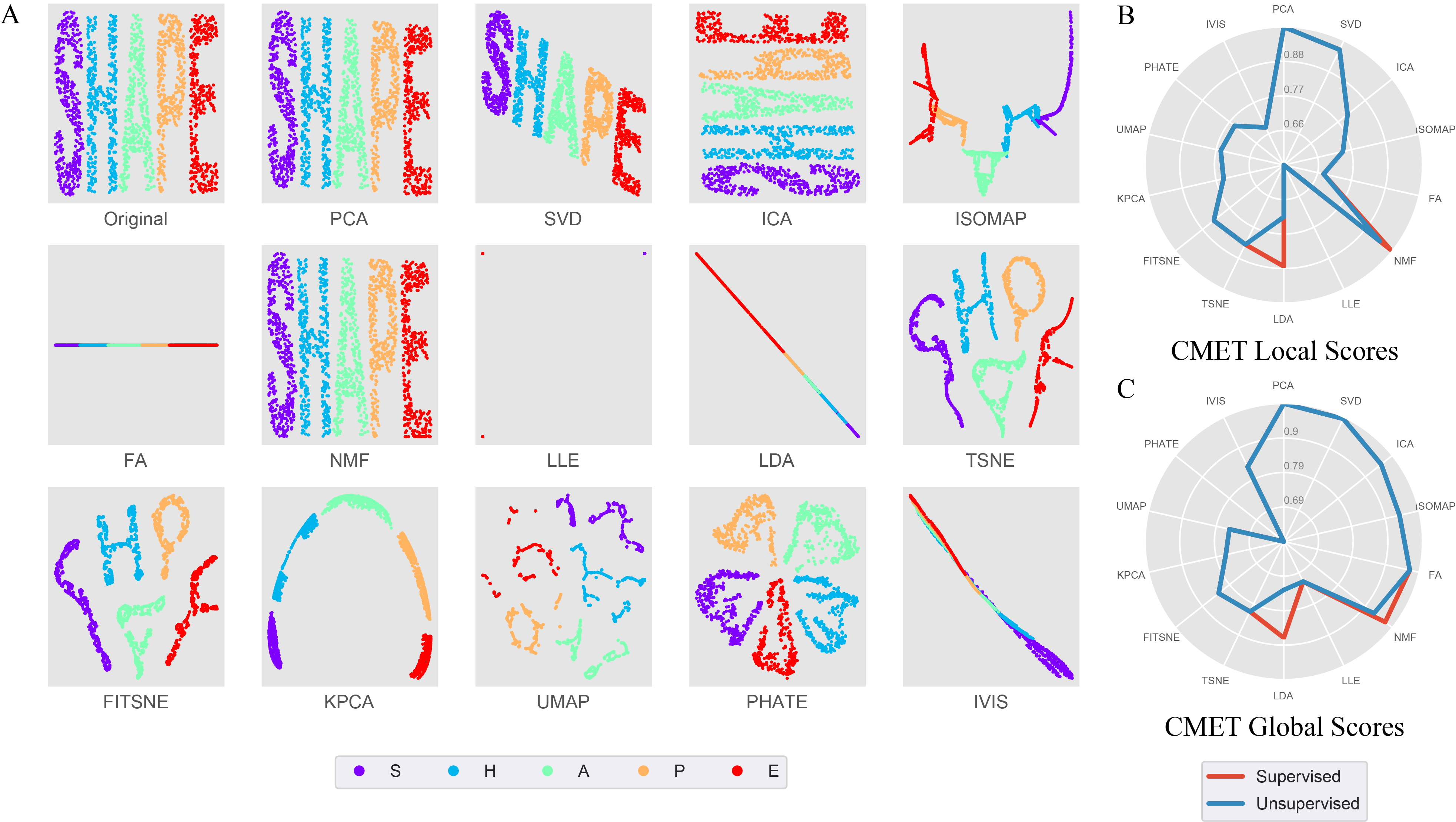}
\caption{Shape Embeddings and CMET Scores}
\label{fig:Shape Embeddings and CMET Scores}
\end{figure}

NMF and PCA have produced almost the original structure with hardly any distortion in local or global shape of the dataset (Figure \ref{fig:Shape Embeddings and CMET Scores}). SVD has rotated the structure, while ICA has made a rotated mirror image like embedding. FITSNE and TSNE have produced very similar embeddings but they differ from the original structure in various ways. All other embeddings are far from resembling the original structure.

Here also, PCA, SVD, NMF have high local CMET scores as their embeddings resemble the original structure to a great extent. FITSNE, TSNE and ICA have scores less than the previous methods but more than others. Global CMET scores are good for PCA, SVD, FA, ISOMAP, NMF, and ICA.

\begin{flushleft}
    \textit{Swiss Roll}
\end{flushleft}

\begin{figure}
\includegraphics[width=\columnwidth]{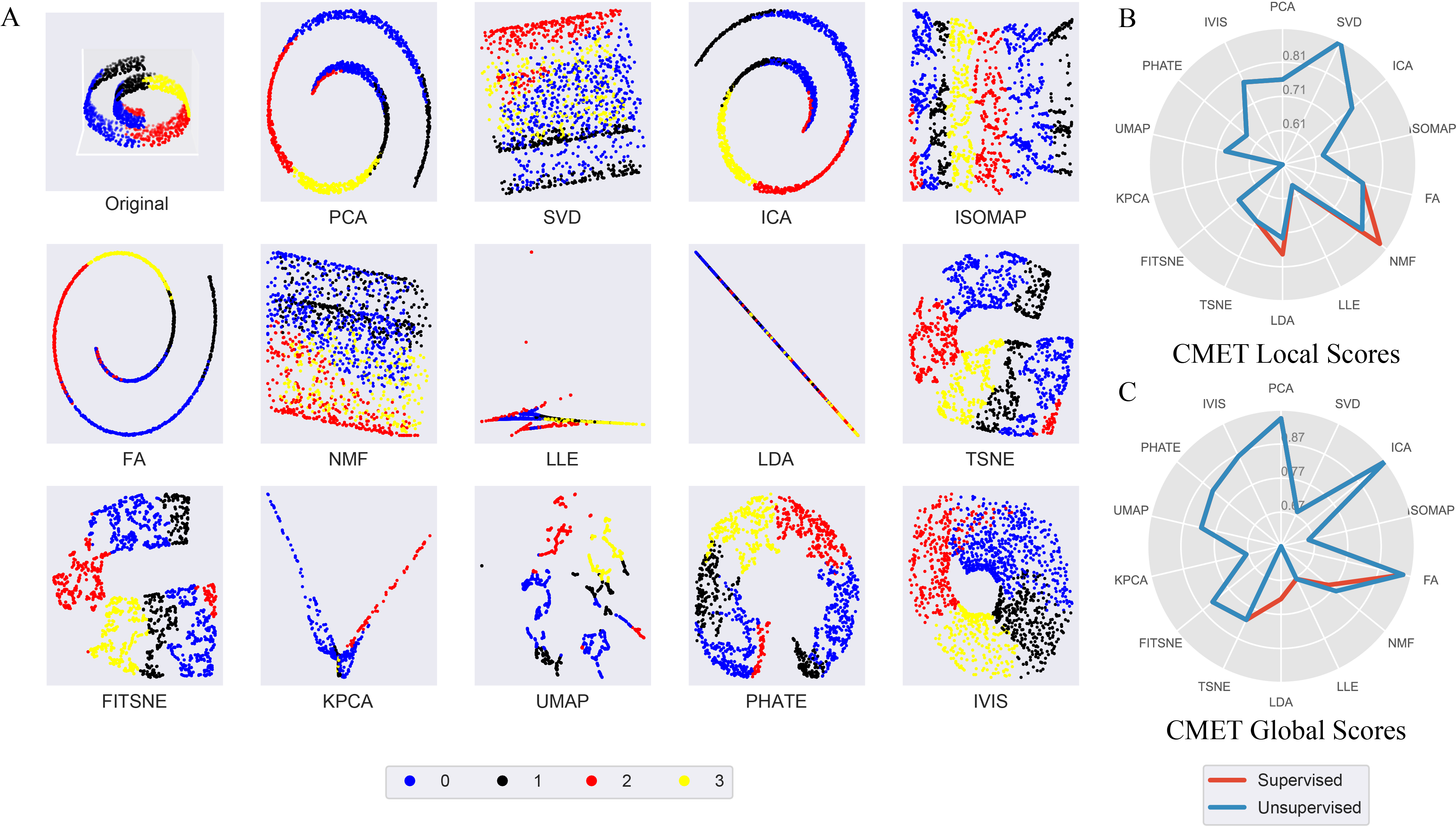}
\caption{SwissRoll Embeddings and CMET Scores}
\label{fig:SwissRoll Embeddings and CMET Scores}
\end{figure}

The original three dimensional data is a spiral structure of four labels. Embeddings of  PCA, ICA, and FA have recovered the spiral structure (Figure \ref{fig:SwissRoll Embeddings and CMET Scores}). As a result, they have outperformed all other DR methods in preserving the global shape of the original data as suggested by CMET global scores of these methods. Band-like local structures are preserved by SVD and NMF resulting in higher  $CMET_L$ values. LDA and LLE have concentrated all the points. 

\subsubsection{Transforming synthetic datasets}
Here, we present results on transformed two dimensional synthetic datasets. Two-dimensional coordinates $(x, y)$ were mapped to a nine-dimensional space by the transformation \cite{IVIS} $$(x, y) \rightarrow (x+y, x-y, xy, x^{2} , y^{2} , x^2y, xy^{2} , x^{3} , y^{3} ).$$ Then they were again reduced to two dimension by the DR methods. This experiment shows how sensitive a DR method is due to transformation. The resulting datasets have been termed as $2-9-2$ along with their original names.

\begin{flushleft}
    \textit{Olympics}
\end{flushleft}

\begin{figure}
\includegraphics[width=\columnwidth]{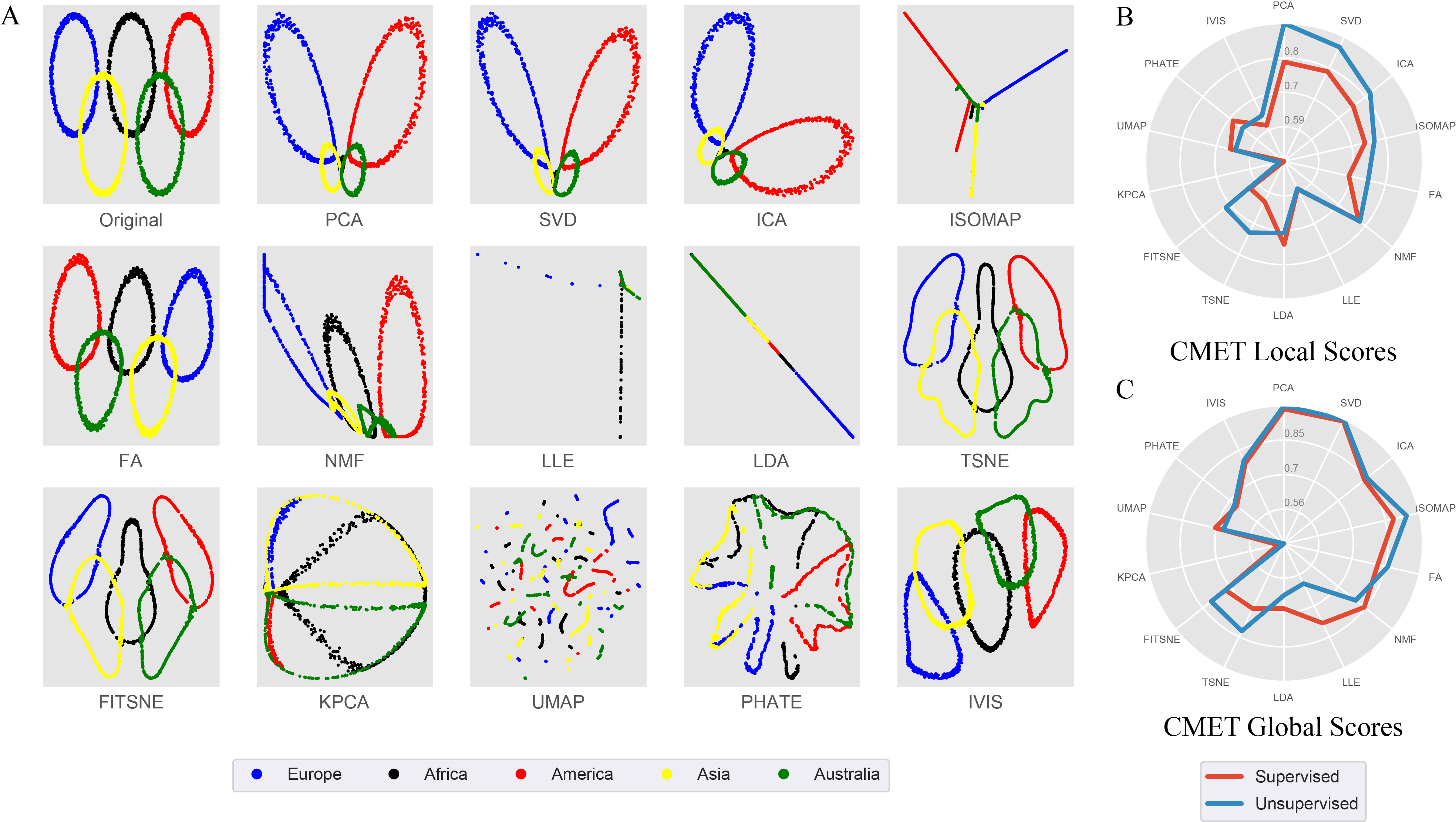}
\caption{Olympics $2-9-2$ Embeddings and CMET Scores}
\label{fig:Olympics 292 Embeddings and CMET Scores}
\end{figure}

PCA, SVD, ICA produced similar embeddings where the red and blue rings are squeezed but looking elliptical, and black ring has become almost invisible. The size of yellow and green rings are also reduced. Both the CMET scores are noted to be high for these embeddings. On the other hand,  TSNE, FITSNE  and IVIS embeddings kept the order of the rings intact. FA embedding kept the rings in the reverse order. All these four embeddings are similar, which is reflected in the $CMET_G$ score. LDA, LLE, KPCA, and PHATE do not have high CMET scores as they have destroyed the structure of the embedding. 

\begin{flushleft}
    \textit{WorldMap}
\end{flushleft}

\begin{figure}
\includegraphics[width=\columnwidth]{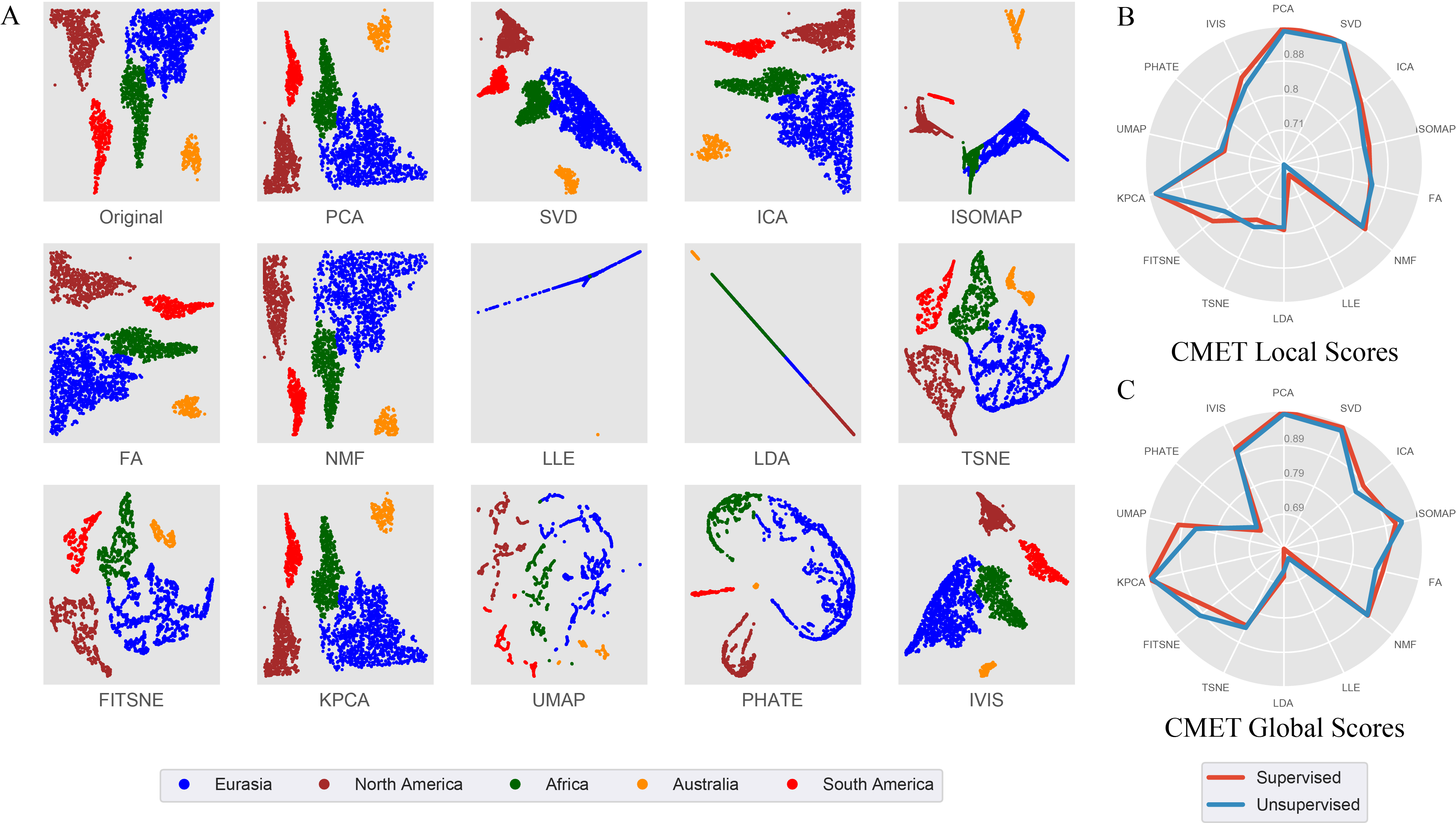}
\caption{WorldMap $2-9-2$ Embeddings and CMET Scores}
\label{fig:World 292 Embeddings and CMET Scores}
\end{figure}
  
Embeddings of PCA, SVD, FA, ICA, and NMF are very similar to the original structure of the data (Figure \ref{fig:World 292 Embeddings and CMET Scores}). All of them have remarkably high local and global scores.  On the other hand,  LLE, PHATE, and LDA have not produced embeddings anyway close to the data.  IVIS also has high CMET scores as it has preserved the clusters. Embeddings of TSNE or FITSNE are a bit different but original structure is reflected in them.

\begin{flushleft}
    \textit{Shape}
\end{flushleft}

\begin{figure}
\includegraphics[width=\columnwidth]{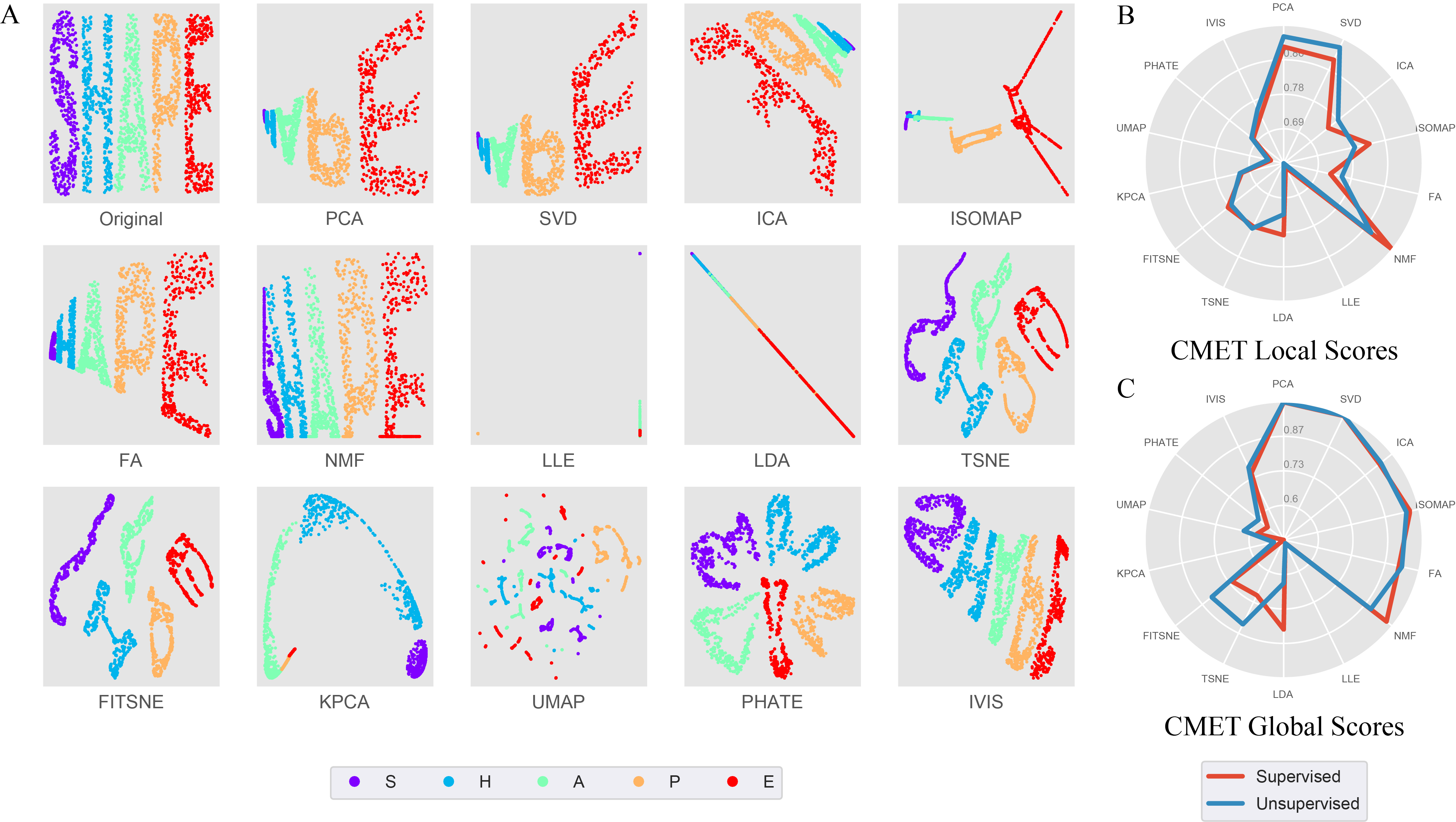}
\caption{Shape $2-9-2$ Embeddings and CMET Scores}
\label{fig:Shape 292 Embeddings and CMET Scores}
\end{figure}

Due to the transformation, points of the letters of the word "SHAPE" have spreaded out decreasingly in the reverse order of the letters (Figure \ref{fig:Shape 292 Embeddings and CMET Scores}). PCA, SVD, and NMF have recovered the structure. They have both higher values of local and global scores.  LLE, KPCA, PHATE, and UMAP have deformed the structure to a great extent. This is also reflected in their respective CMET scores.

\subsubsection{Biological datasets}

\begin{flushleft}
    \textit{Jurkat}
\end{flushleft}

\begin{figure}
\includegraphics[width=\columnwidth]{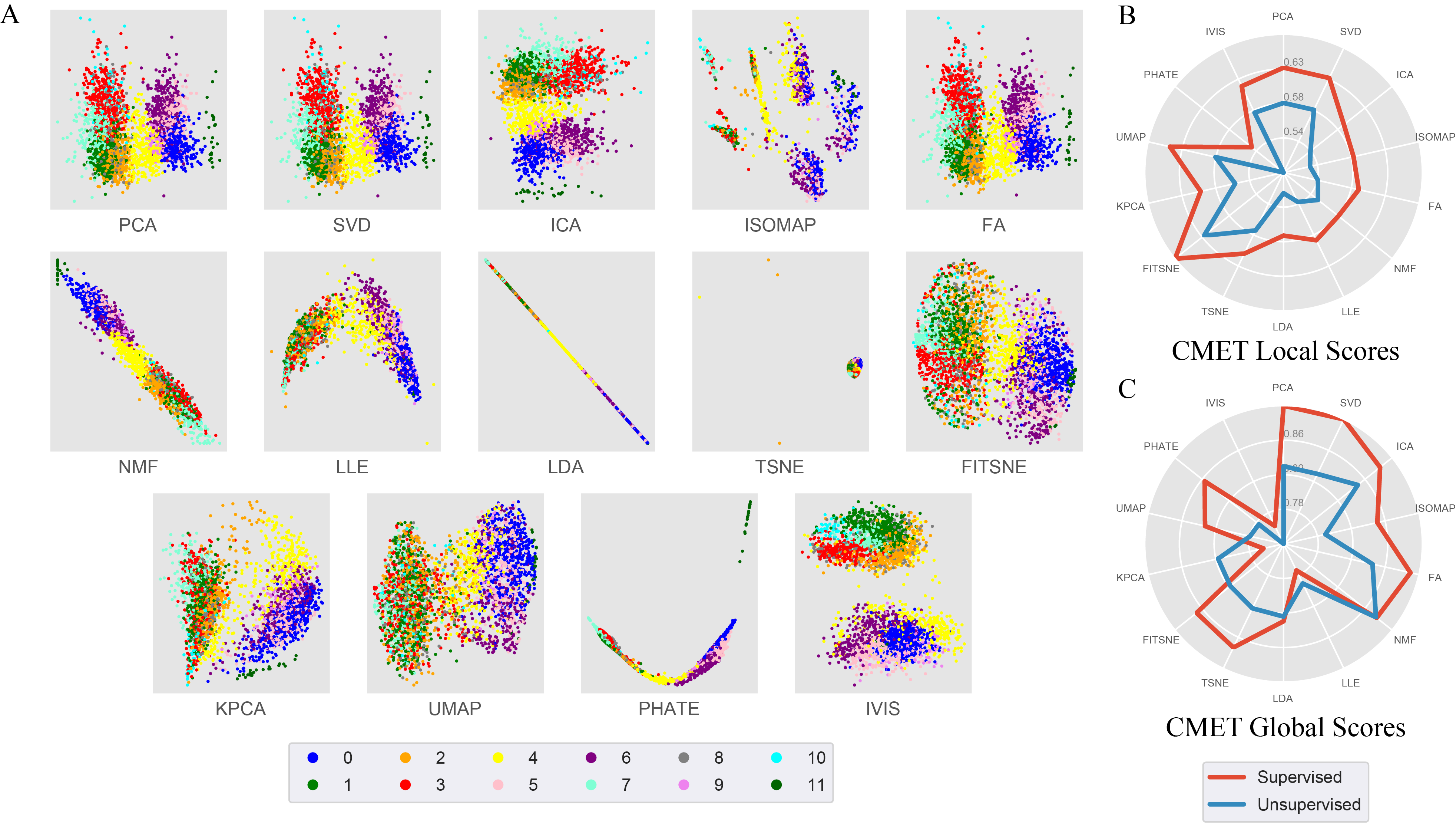}
\caption{Jurkat Embeddings and CMET Scores}
\label{fig:Jurkat Embeddings and CMET Scores}
\end{figure}

Jurkat data, derived from the human T lymphocyte Jurkat cell line, captures gene expression or proteomic profiles relevant to T-cell signaling and immune responses. This dataset has been preprocessed by the standard pipeline as recommended by Scanpy \cite{scanpy}. PCA, SVD, ICA, FA, and IVIS have produced alike embeddings (Figure \ref{fig:Jurkat Embeddings and CMET Scores}). They have close $CMET_L$ values as well as $CMET_G$ values. However, embeddings of FITSNE and UMAP have exceeded the previous ones with respect to $CMET_L$ scores. Global scores of PCA, SVD, ICA, FA, and NMF are better than that of FITSNE and UMAP. LLE, NMF, and LDA have significantly lower local CMET scores than others. Supervised and unsupervised CMET scores have a similar trend.

\begin{flushleft}
    \textit{Zeisel}
\end{flushleft}

\begin{figure}
\includegraphics[width=\columnwidth]{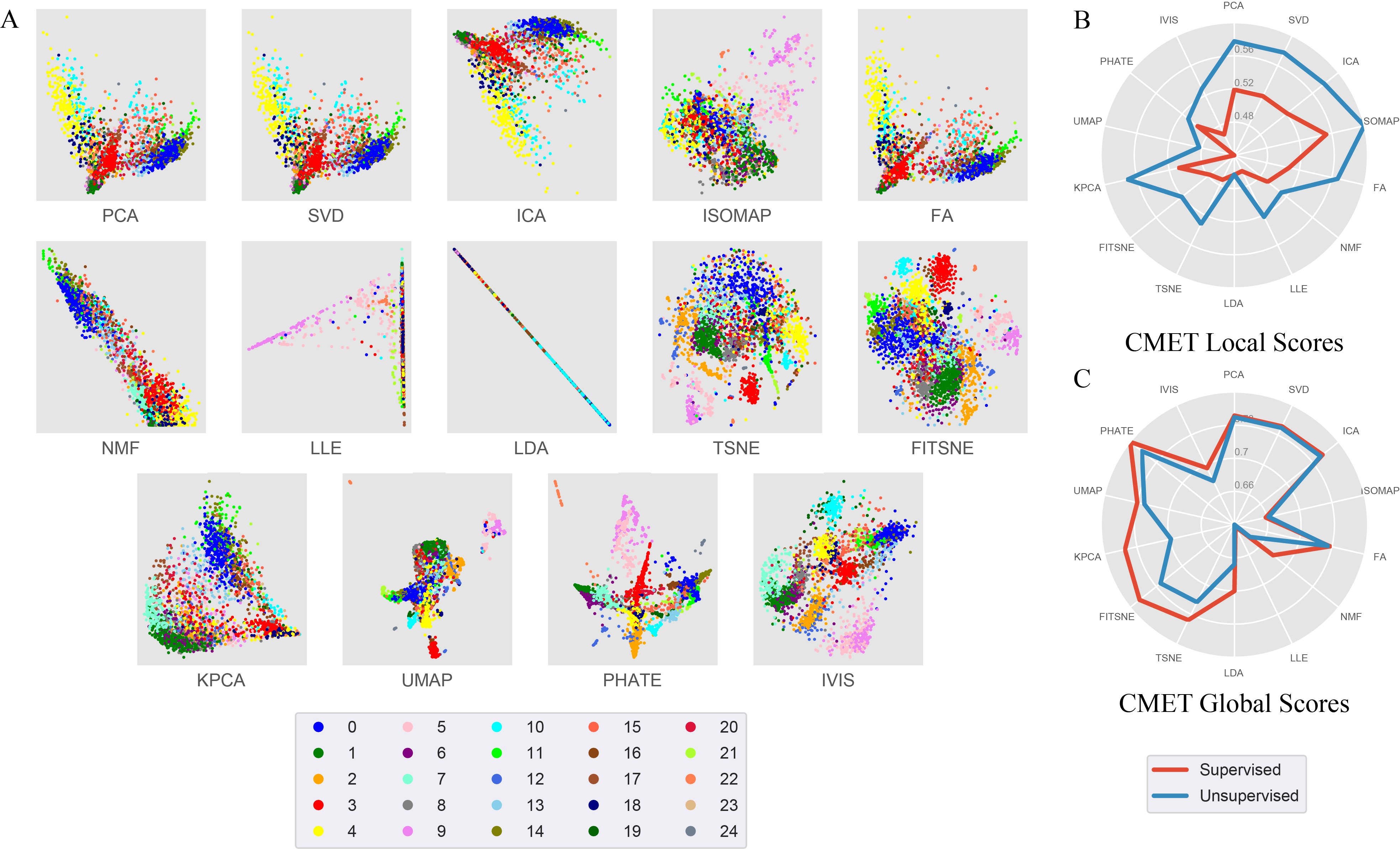}
\caption{Zeisel Embeddings and CMET Scores}
\label{fig:Zeisel Embeddings and CMET Scores}
\end{figure}

Zeisel data consists of single-cell RNA sequencing profiles from mouse brain cells, providing a comprehensive view of cellular diversity and gene expression in the nervous system. Similar to Jurkat dataset,  Zeisel dataset has also been preprocessed by the standard pipeline as recommended by Scanpy. 
PCA, SVD, ICA, KPCA, and FA have produced similar embeddings (Figure \ref{fig:Zeisel Embeddings and CMET Scores}). As a result, all of them have similar  $CMET_L$ as well as similar $CMET_G$ values. ISOMAP has the highest local score.  In case of global shape preservation, PHATE has the highest  $CMET_G$ value. UMAP, TSNE, and FITSNE also have high $CMET_G$ values. Methods like LLE and LDA have produced embeddings where data points have been concentrated, leaving no notion of local or global structure, and yielding low CMET scores. Here also, supervised and unsupervised CMET scores have a similar trend for both local and global scores.

\subsubsection{Image datasets}

\begin{flushleft}
    \textit{MNIST}
\end{flushleft}

\begin{figure}
\includegraphics[width=\columnwidth]{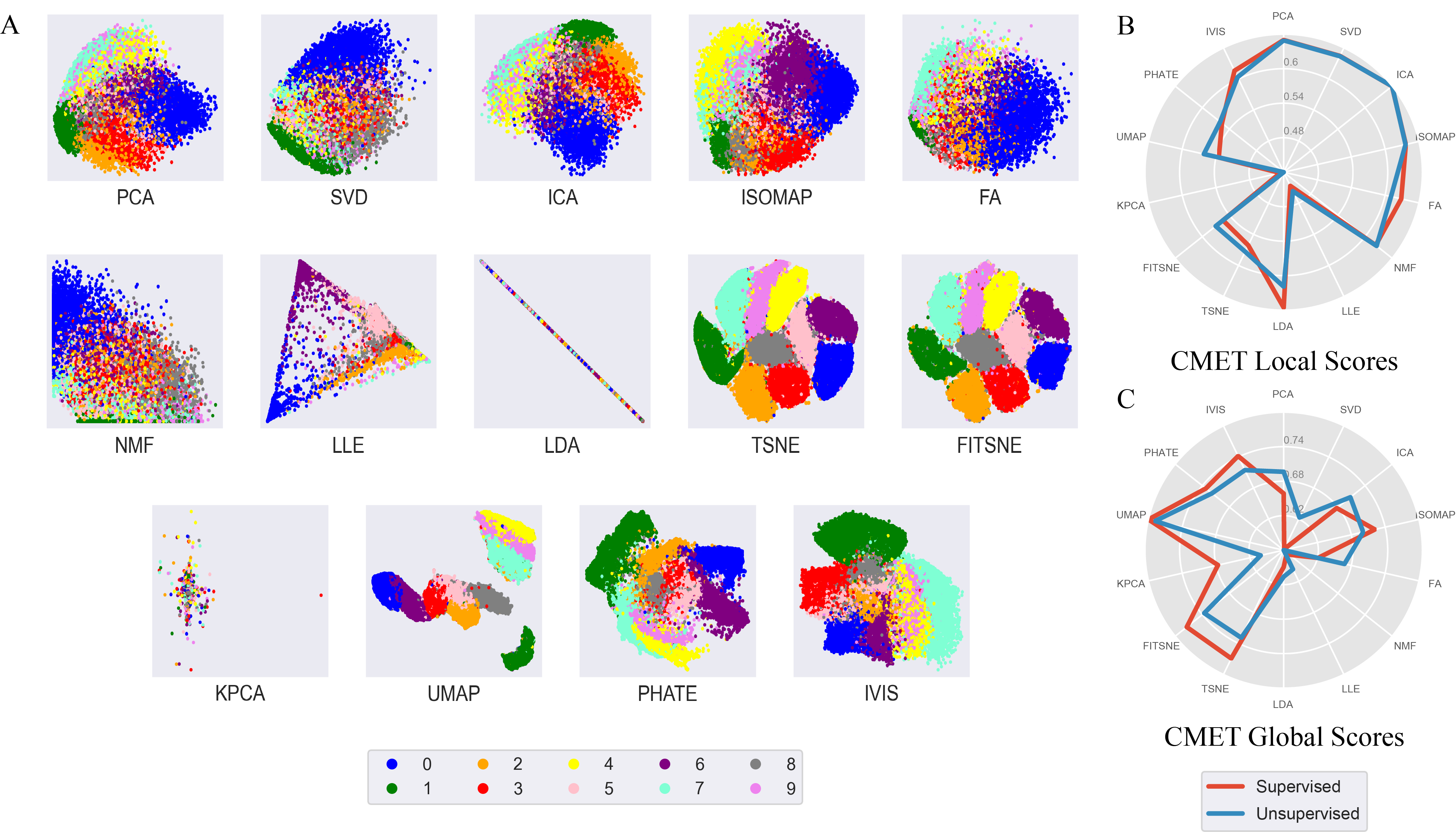}
\caption{MNIST embeddings and CMET Scores}
\label{fig:MNIST embeddings and CMET Scores}
\end{figure}

MNIST (Modified National Institute of Standards and Technology database) is a well known database of hand written digits consisting of $60,000$ data points each of which is an image of one of the ten digits. As often illegible hand writing leads to difficulty in recognizing two similar looking digits, it is expected that Agglomerative clustering will be unable to make ten clusters perfectly as the same as the ten labels. However, similar pattern of the supervised and unsupervised lines for local and global CMET scores (Figure \ref{fig:MNIST embeddings and CMET Scores}) asserts that the clusters are enough separated with some overlapping. Thus, two dimensional embeddings are expected to place all the clusters almost distinctly. As visual differentiation is not possible for data of higher dimensionality, this approach enables us to make some judgement on the embeddings. Because of its large size, we could not calculate any other metric with our available computational facility, whereas CMET took significantly less amount of time to calculate both the local scores. Plot of the embeddings insists that TSNE, FITSNE, UMAP, PHATE, and IVIS have separated the clusters to a great extent, whereas PCA, SVD, ICA, ISOMAP, NMF, and FA have mingled up all the clusters, although, except FA, all the last five DR methods have produced similar embeddings. Global CMET scores also suggest that TSNE, FITSNE, UMAP, PHATE, and IVIS have higher $CMET_G$ value than most of the other methods.  PCA, SVD, ICA, ISOMAP, NMF, and FA have significantly lower score than the other methods but since they produced similar embeddings, they have their scores vary close to each other, bound in a very small interval. Embeddings of LDA and LLE show that these methods have concentrated all the points together,  resulting in very low $CMET_L$ score. However, instead of producing separated clusters,  TSNE, FITSNE, and UMAP do not result in good $CMET_L$ score.  As pointed out previously, PCA and SVD have produced similar embeddings causing to have similar higher score of  $CMET_L$ together.

\begin{flushleft}
    \textit{FMNIST}
\end{flushleft}

\begin{figure}
\includegraphics[width=\columnwidth]{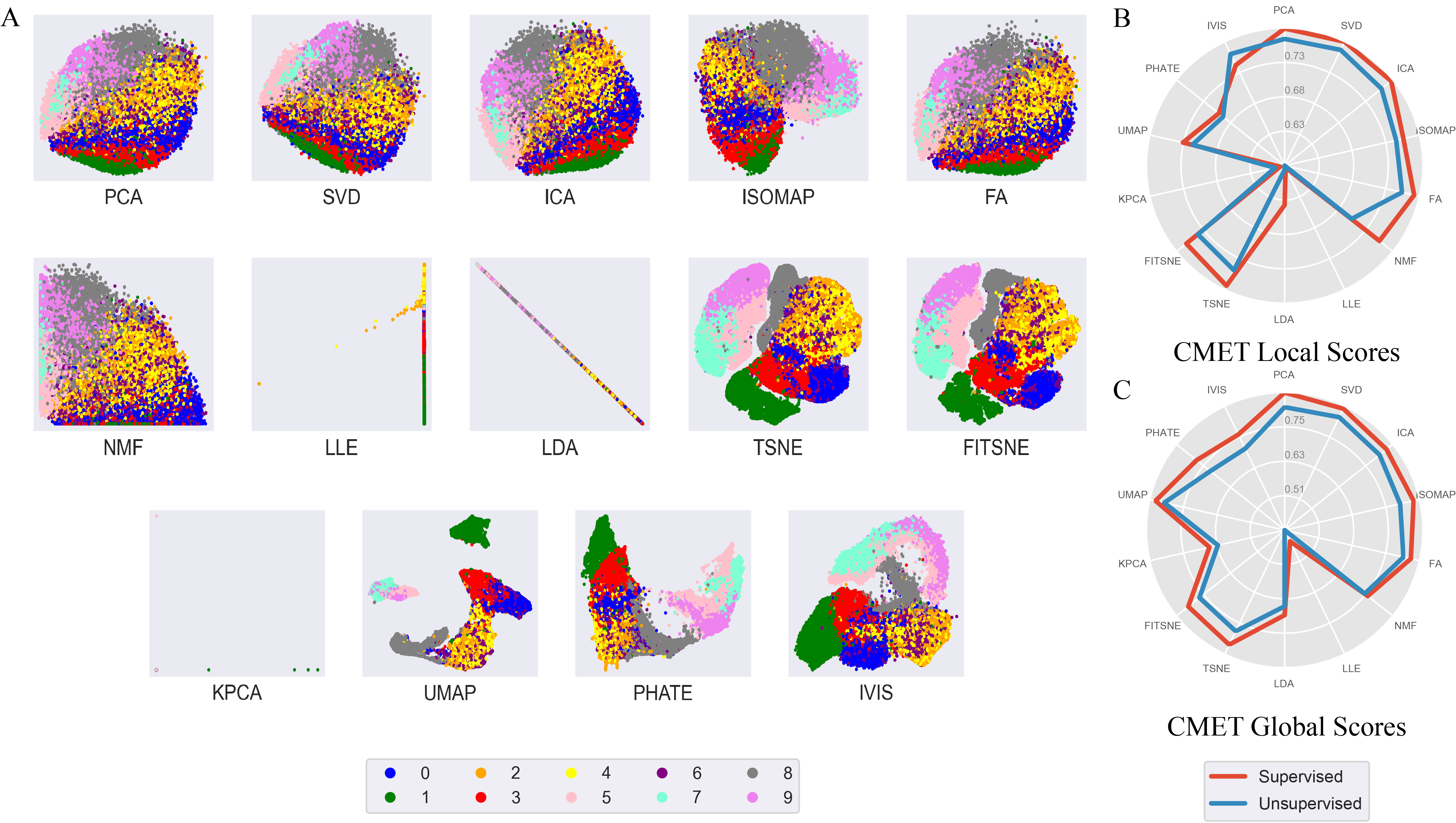}
\caption{FMNIST embeddings and CMET Scores}
\label{fig:FMNIST embeddings and CMET Scores}
\end{figure}

In $2017$, Fashion-Mnist (FMnist), another highly well-liked image dataset that is frequently utilized in machine learning,  was developed. Each of the $60,000$ training samples in this dataset, which is made up of $7,000$ images per category of Zalando article images, is a $28 \times 28$ grayscale image of a fashion product from one of ten categories. The data comprises images of a T-shirt, trousers, a coat, a bag, pullover. The CMET spider-plots (Figure \ref{fig:FMNIST embeddings and CMET Scores}) are well-assuring to consider the labels to be separated enough. Embeddings produced by PCA, SVD, ICA, ISOMAP, FA, NMF, TSNE, FITSNE, and IVIS are very similar, and all of them have very high $CMET_L$ scores, as well as $CMET_G$ scores. On the other hand, LLE, KPCA, and LDA have produced embeddings that are way different, having no clue of clusters. Thus, both of the CMET scores, among others, are very low for these three methods.

\subsection{Sensitivity of the hyper-parameter}
\label{subsec:Sensitivity of the Hyper-parameter}

CMET is designed to assess the extent of the local and global shape of a dataset preserved under a certain transformation. Methodologically, it relies upon the underlying unsupervised method of clustering that is being used. In this study, we have used Agglomerative clustering, as it is very flexible to use and it does not need any strong prior knowledge about the number of clusters of the original dataset. It can be used to get the dataset clustered to any number, which is a hyper-parameter of this clustering algorithm. Thus, the number of clusters becomes a hyper-parameter for CMET involving Agglomerative clustering, and plays a critical role in deciding CMET scores. As pointed out earlier, CMET uses the clustering method only to partition the dataset into a specified number of closely located groups so that the local and global structure of the data is revealed in a particular way.  However, when there is no notion about the number of potential clusters, it becomes challenging to arrive at a single value for this hyper-parameter.  In this section, an impact analysis is presented around the influence of the hyper-parameter on CMET scores.  

More often than not labeled data give a clear notion about the number of clusters, which is reflected in the similar patterns of supervised and unsupervised CMET values of the embedding shown in the above subsection. Unsupervised approach is more advisible to use even in the presence of data labels, if the labels are unable to yield a sensible clustering in the data. On the other hand, it is also not desirable for the hyper-parameter to be sensitive, as it might make CMET scores unstable. 

\begin{figure}
\includegraphics[width=\columnwidth]{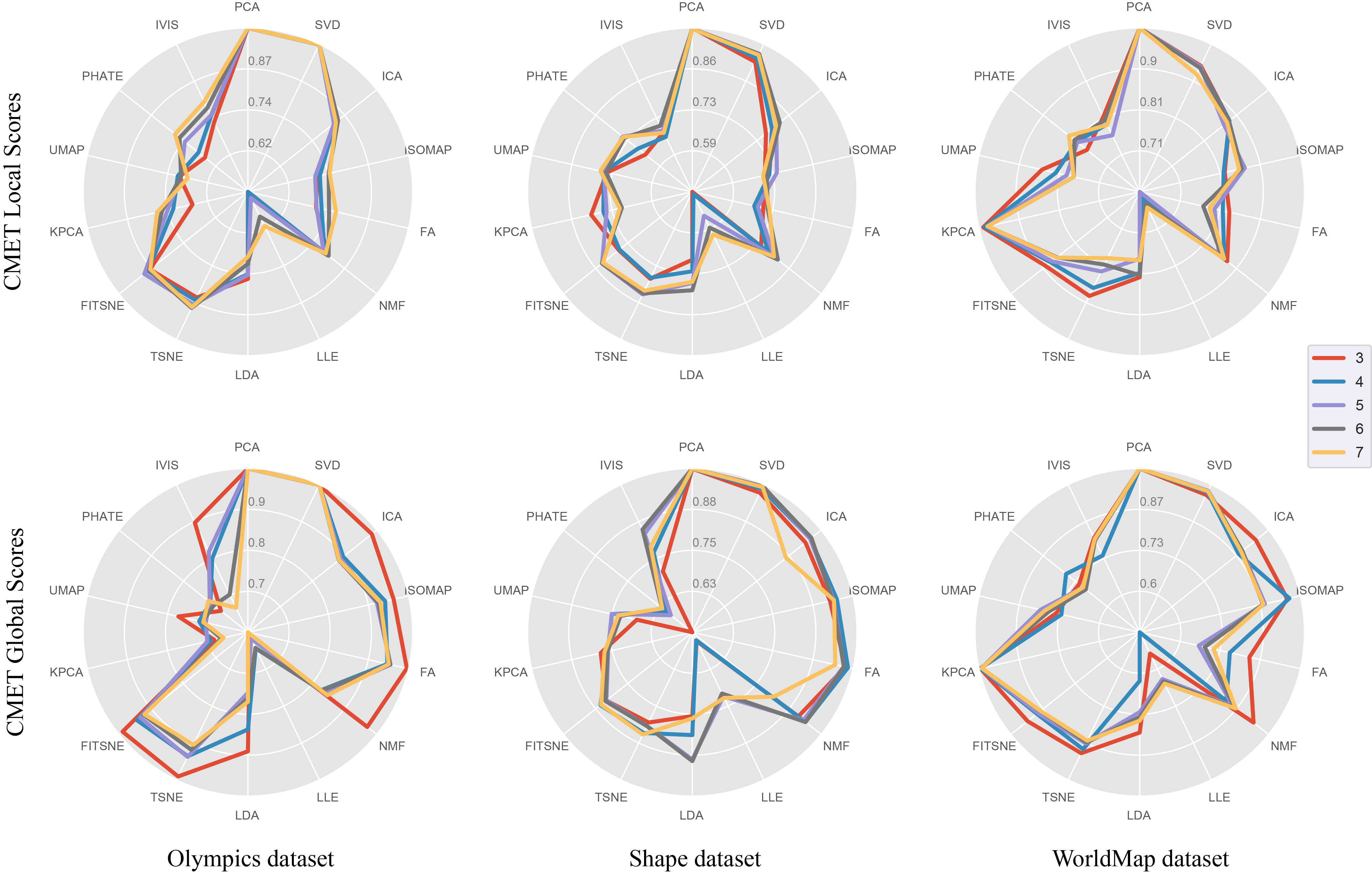}
\caption{Sensitivity of CMET parameter for two-dimensional synthetic datasets}
\label{fig:2d sensitivity}
\end{figure}

\begin{figure}
\includegraphics[width=\columnwidth]{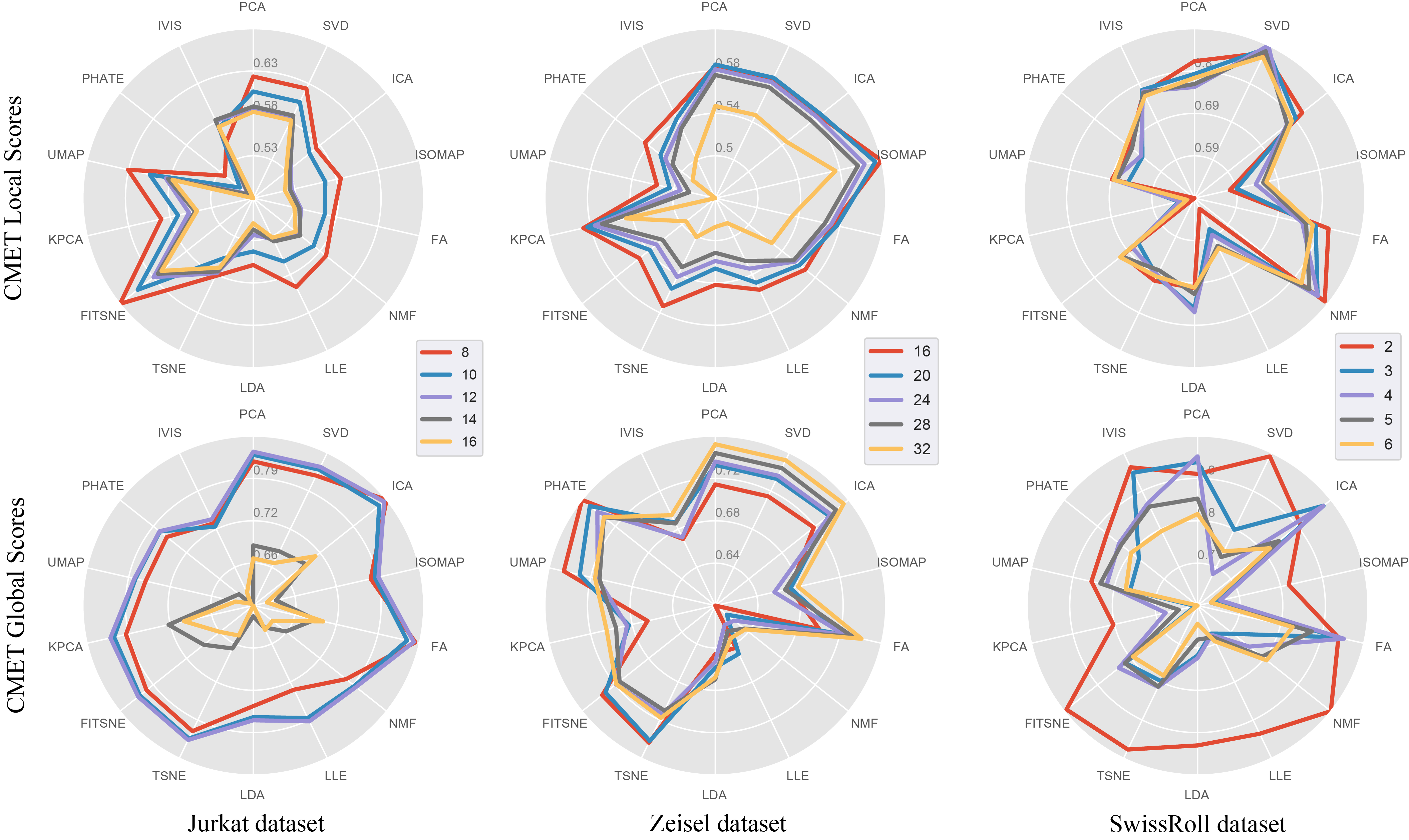}
\caption{Sensitivity of CMET parameter for SwissRoll and Biological datasets}
\label{fig:3d & bio data sensitivity}
\end{figure}

The following experiment has been carried out to establish the stability of the CMET scores against a possible set of hyper-parameters. In the last section, local and global CMET scores have been calculated by taking the number of clusters being equal to the number of labels for each dataset. Now, we present CMET scores for five different equidistant values of the hyper-parameter, the middle-most being the number of labels. Experimental results are summarised by the spider plots shown in Figures \ref{fig:2d sensitivity}-\ref{fig:3d & bio data sensitivity}.

$Olympics$ dataset consists of five rings colored differently in the scatter plot representing the five labels it has. While presenting the original results on the performance of all the methods considered here, $k=5$ was used. Now, both the CMET scores for $k=3,4,5,6,7$ are depicted with the help of a spider plot (Figure \ref{fig:2d sensitivity}). The local and global sensitivity plots show that $CMET_L$ increases and $CMET_G$ decreases, for most of the DR methods, as $k$ increases. Except for IVIS in the global case, the pattern remains unchanged for the remaining DR methods across all values of $k$ for both local and global scores.

The spider plot for $Shape$ data shows that local and global scores generally increase with increasing $k$. Here, the values of $CMET_L$ follow an increasing pattern for KPCA, and for $k = 3, 4, 5$, and $6$, the pattern of the global sensitivity is all the same. For LDA, LLE, and ICA,  $CMET_G$ values are a bit noisy.

$WorldMap$ dataset resembles the political map of the world, which consists of five labels for five continents. Here also, apart from PHATE and ISOMAP in the local case, and LLE, UMAP and PHATE in the global case,  $CMET_L$ and $CMET_G$ tend to decrease as $k$ increases. For $k = 3,4$ and $5$, $CMET_L$ and $CMET_G$ overlap for all the DR methods. It may be mention here that the trend of the scores are preserved for all the synthetic datasets.

$SwissRoll$ dataset is a very popular three-dimensional data with four labels looking like a rolled curvy plane. Here, $k=2,3,4,5,$ and $6$ have been considered for $k$ (Figure \ref{fig:3d & bio data sensitivity}).  The local sensitivity plot is very stable with minimal anomaly in the pattern, which is more or less increasing in nature. However, particularly for $k=$2, $CMET_G$ values are very different from others. Actually, by definition of $CMET_G$, if the dataset has exactly one cluster, $CMET_G$ value will be $1$. As two clusters also do not give much information about the global structure, $CMET_G$ values tend to be high for most of the DR methods. For other values of $k$, the curves overlap and create a stable pattern with some ignorable noise.

Since $Jurkat$ data has $12$ labels, unlike the synthetic data, here $CMET_L$ and $CMET_G$ values are shown for $k=8,10,12,14$, and $16$, i.e., for five consecutive even numbers starting from $8$ (Figure \ref{fig:3d & bio data sensitivity}). Only a slight change is observed in the value of $CMET_L$ for $k=8$. For $k=12, 14,$ and $16$, the lines are overlapping. In most of the cases, for $ k=8$ and $10$, $CMET_L$ values are higher than that for values of $k=12,14$, and $16$, but the order remains the same, which is followed by the others. An interesting phenomenon is found in the global sensitivity plot. $CMET_G$ values almost overlap for $k= 8, 10$, and $12$, but there is a significant lowering in the values of $CMET_G$ for $k=14$ and $16$, but the pattern for these two cases agrees, and also the same as the pattern created by the three others.

$Zeisel$ dataset has $24$ labels. Since it is the largest of all the previously discussed datasets, five values of $k$ have been taken to be widely spread. No fluctuation of $CMET_L$ values is observed for any of the DR methods. A strictly decreasing trend with $k$ value is followed. In the global sensitivity plot, no overlap is found, but the pattern very objectively indicates the order.

\subsection{Comparison with some other Measures}
\label{subsec:Comparison with Other Metrics}

In this section, a comparison of the performance of CMET with three other metrics, viz., trustworthiness, continuity, and LCMC, is presented. With our limited computational resources, we could not calculate these three metrics for large datasets, like MNIST and F-MNIST. On the contrary, CMET took significantly less time to come up with local and global scores for these datasets. In this way, the scalability of CMET is established.

\begin{figure}
\includegraphics[width=\columnwidth]{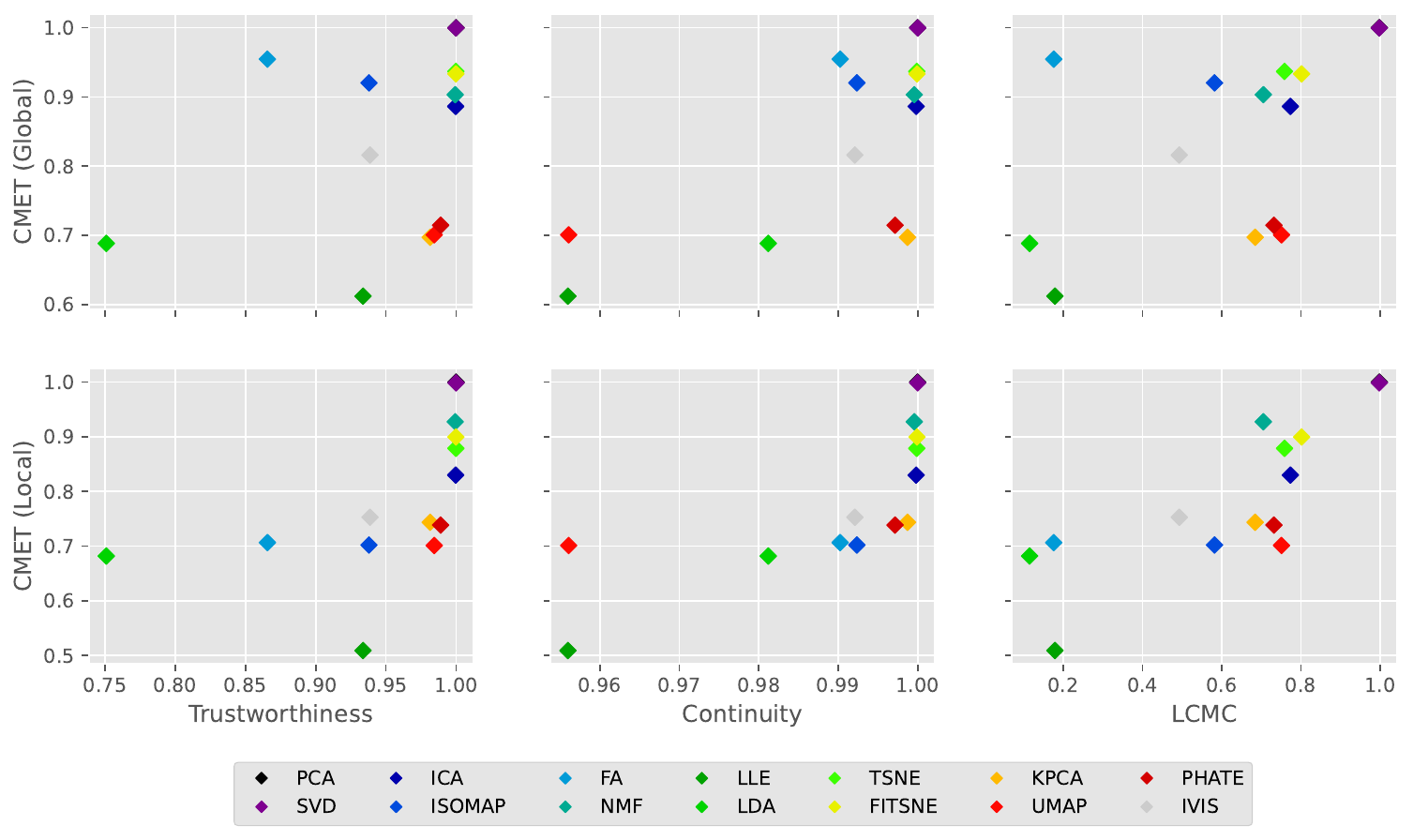}
\caption{Comparison of CMET scores with the state-of-the-art methods for Olympics dataset}
\label{fig:Olympics Comparison Plots}
\end{figure}
\begin{figure}
\includegraphics[width=\columnwidth]{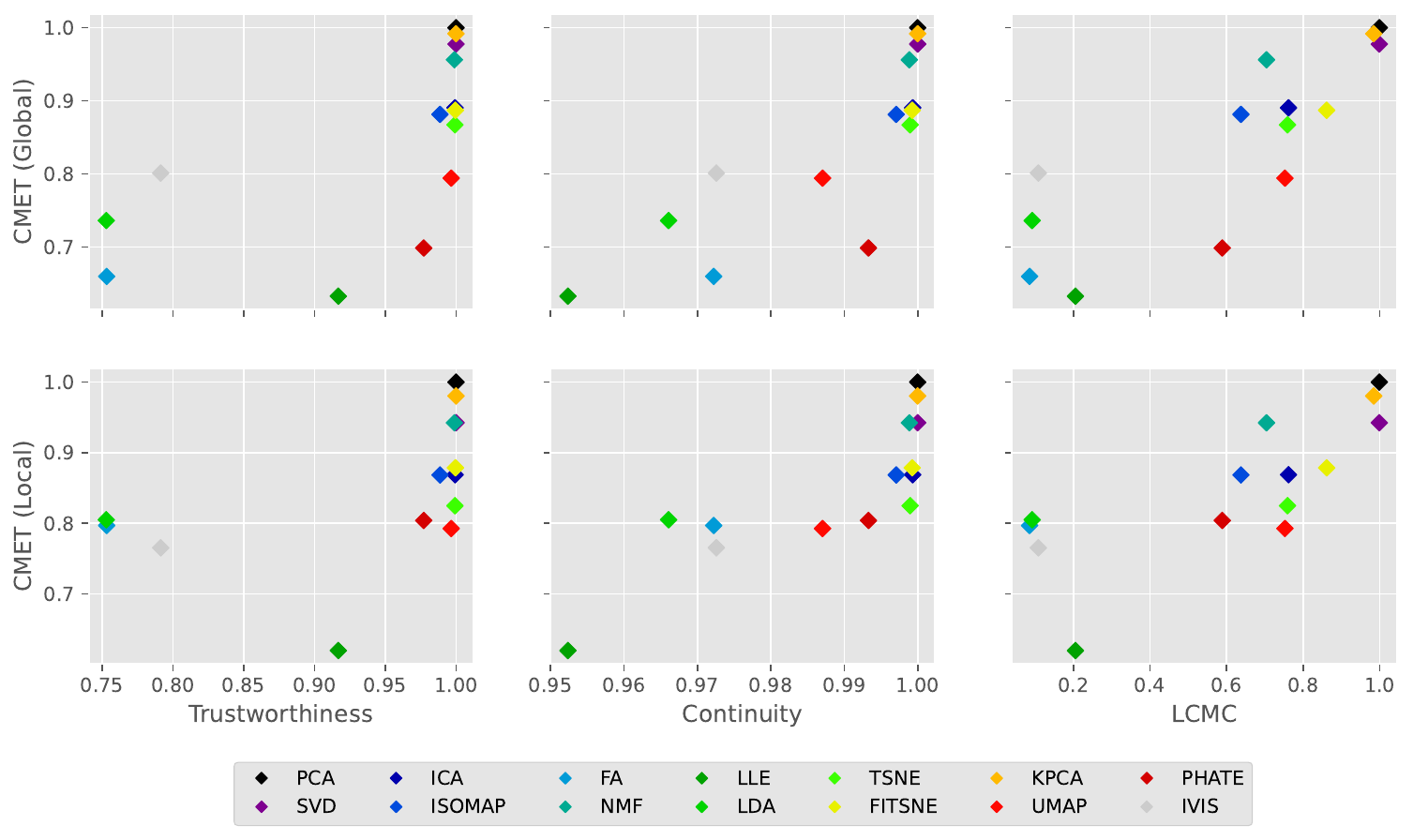}
\caption{Comparison of CMET scores with the state-of-the-art methods for WorldMap dataset}
\label{fig:World Comparison Plots}
\end{figure}
\begin{figure}
\includegraphics[width=\columnwidth]{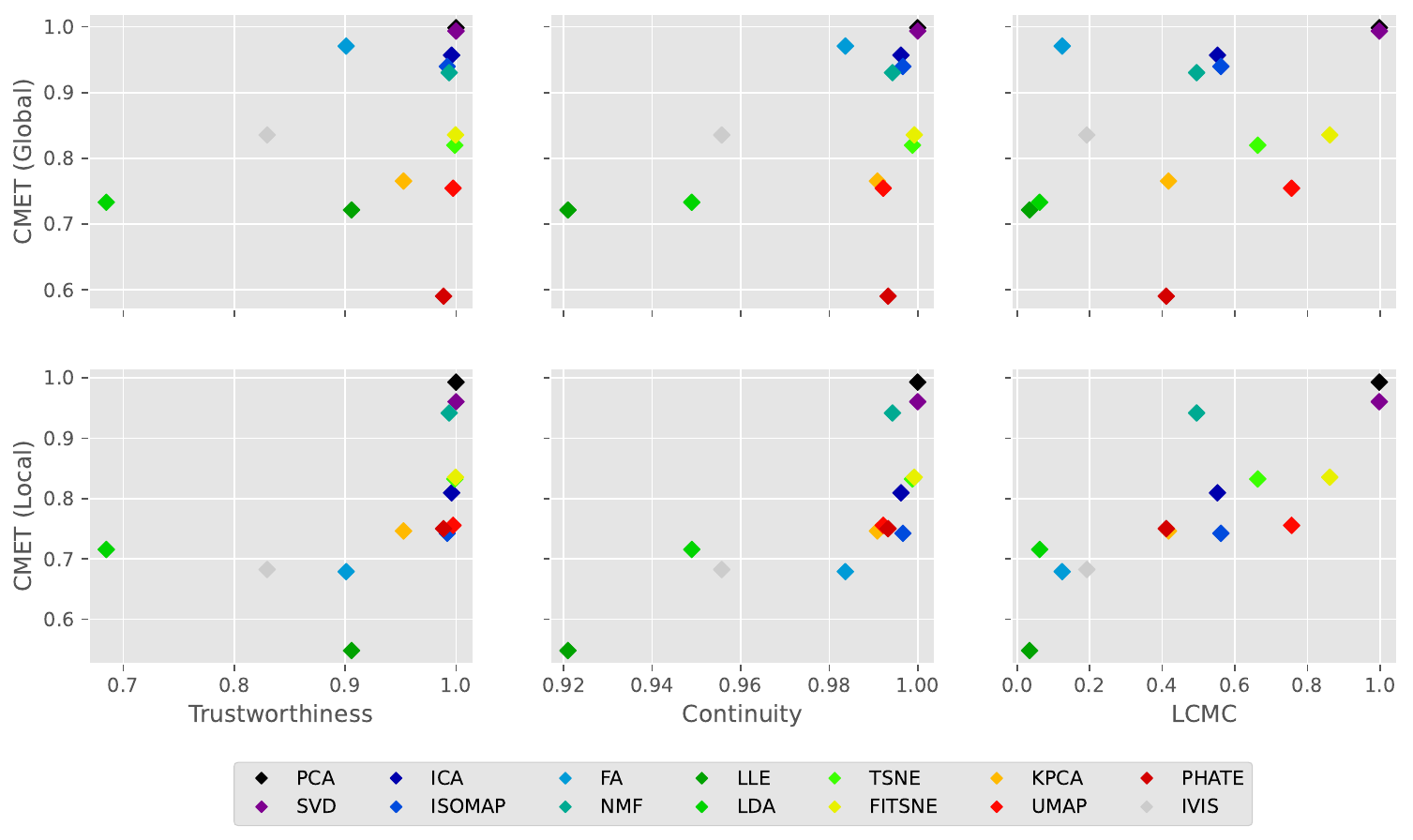}
\caption{Comparison of CMET scores with the state-of-the-art methods for Shape dataset}
\label{fig:Shape Comparison Plots}
\end{figure}

All the four methods, including CMET, take values in $[0,1]$, where higher values imply better shape preservation for all of the metrics. $CMET_L$ verses Trustworthiness plots for all the two dimensional data sets (Figure \ref{fig:Olympics Comparison Plots}-\ref{fig:Shape Comparison Plots}) show that Trustworthiness values for six methods (PCA, SVD, ICA, TSNE, FITSNE, and NMF), suggest that all these DR methods are equally performing well but CMET has given an order of these methods. It indicates that CMET can distinguish the DR methods based on their performance. However, for these data sets, $CMET_G$ verses Trustworthiness plots show that PHATE and UMAP have higher values, but their embeddings are not very similar to the original data. The same scenario is observed in the continuity plot. However, the ordering of performances of DR methods in local shape preservation is very similar for both continuity and $CMET_L$. LCMC values are spread over the entire range of $(0,1)$. However, the order of DR methods matches well.

From comparison plots of $WorldMap$ data (Figure \ref{fig:World Comparison Plots}), we note that in the case of preservation of the local shape, all metrics declare LLE to be the least effective while PCA to be the most. Trustworthiness declares many DR methods to have a score exactly equal to one. However, CMET makes a meaningful order of them. The continuity plot gives coherence among the orders of the DR methods. Unlike trustworthiness, the LCMC is capable of finely distinguishing the order of performance, producing distinct scores corresponding to the DR methods.

Here we observe that scores of trustworthiness vary between $0.65$ and $1$, whereas score of continuity varies in $[0.92,1]$. Both of the CMET scores lie in [$0.6-1$]. Again, scores of $CMET_L$ and Continuity match a lot. Trustworthiness tends to declare several DR methods to have a score of $1$, where they differ in their embeddings. CMET has successfully differentiated the minute differences. However, in this dataset too, the least effective DR method is LLE, and the most effective one is PCA, as suggested by all the measures under consideration.

\begin{figure}
\includegraphics[width=\columnwidth]{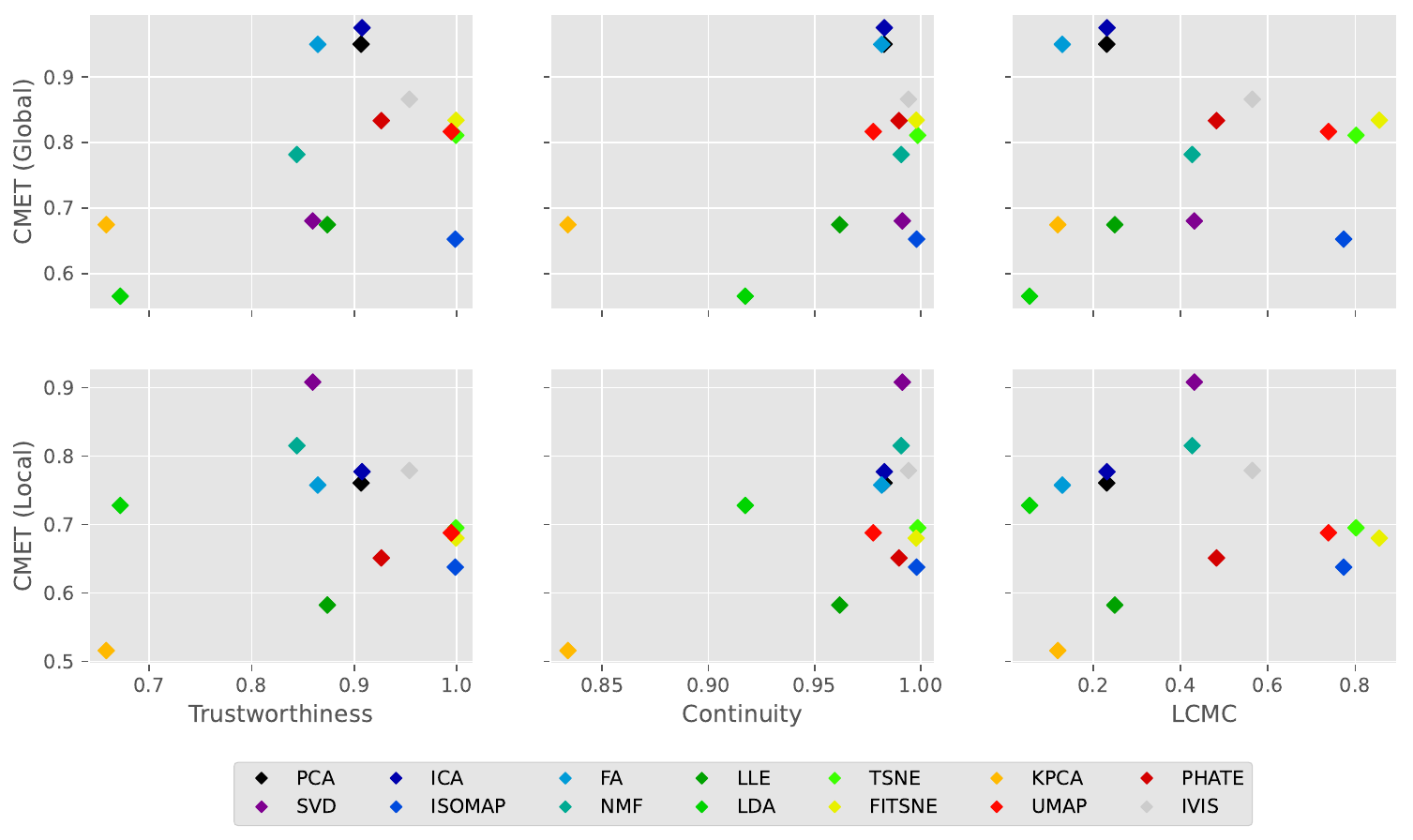}
\caption{Comparison of CMET scores with the state-of-the-art methods for SwissRoll dataset}
\label{fig:SwissRoll Comparison Plots}
\end{figure}

Besides, FITSNE is the least efficient DR method suggested by all the measures. Continuity is prone to declare all the methods to be more or less equally effective, except for a very few. However, the embeddings are suggestive of accepting only a few to be the best, which synchronizes with the result of CMET. LCMC supports PHATE, TSNE, and UMAP to do better in local shape preservation than many others, where CMET contradicts it, as the embeddings are not at all similar to the original embedding of $SwissRoll$.

\begin{figure}
\includegraphics[width=\columnwidth]{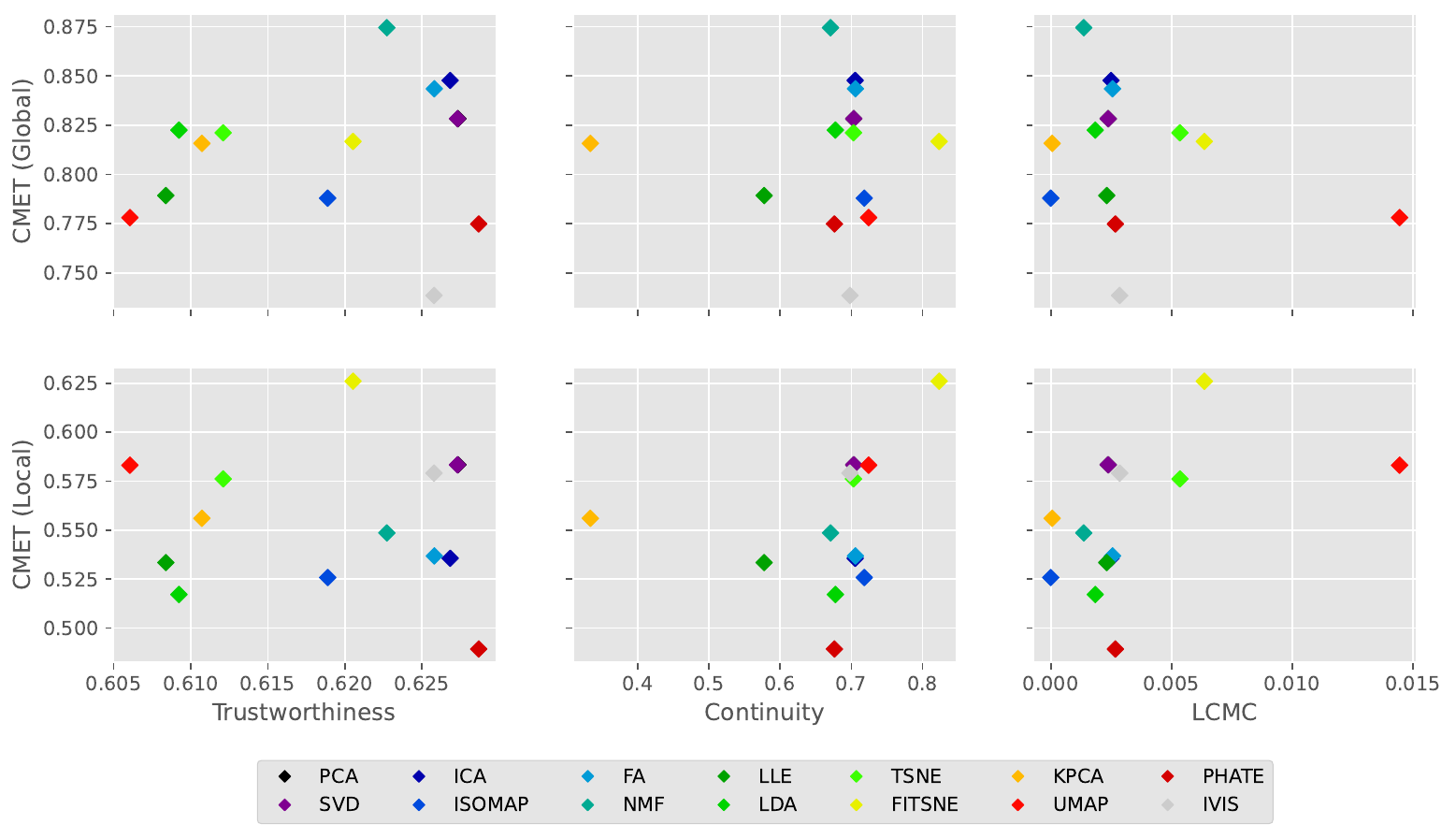}
\caption{Comparison of CMET scores with the state-of-the-art methods for Jurkat dataset}
\label{fig:Jurkat Comparison Plots}
\end{figure}

 Continuity scores are concentrated around $0.7$. Thus, relative performance is not so obvious, whereas CMET distinguishes the DR methods finely. PHATE is declared to be the least while FITSNE is the most effective by all the measures. LCMC scores are very low for all of the DR methods considered here. For data with high dimensionality, LCMC has yielded very less significant results for comparison among DR methods.

\begin{figure}
\includegraphics[width=\columnwidth]{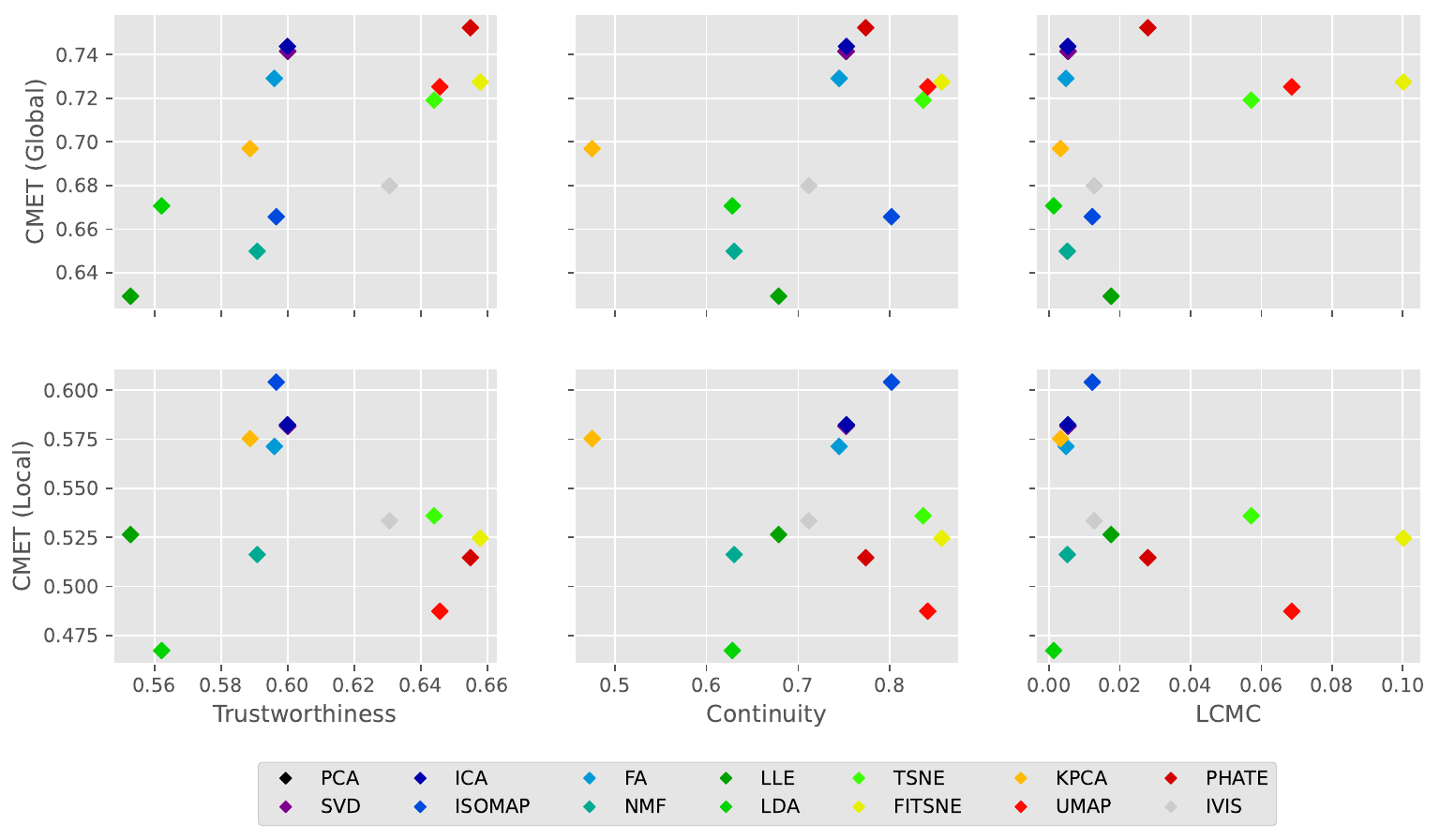}
\caption{Comparison of CMET scores with the state-of-the-art methods for Zeisel dataset}
\label{fig:Zeisel Comparison Plots}
\end{figure}

For $Zeisel$ data, except for CMET, all other measures suggest that FITSNE, TSNE, and PHATE are very effective. $CMET_L$ suggests that ISOMAP, ICA, and KPCA are good at local shape preservation, while as per $CMET_G$ PHATE, UMAP, PCA, and SVD are good at global shape preservation. However, LLE and LDA are pointed out to perform worse than most of the other methods. Moreover, $CMET_G$ has similar results to other metrics. Thus, the methods that are considered to be good at preserving shape by other measures are objectively said to be good at preserving the global shape of this data. Interestingly, the drawback, being subjective to decide whether a DR method is good at preserving local or global shape, trustworthiness, continuity, or LCMC, is mitigated by CMET, which can offer separate scores for both local and global shape preservation capabilities of DR methods.


\section{Discussion and conclusion}
\label{sec:DisCon}

Data transformation related to the projection of the data into a different dimensionality is an essential task for analyzing complex data. In this study, we mainly dealt with dimensionality reduction, which is very helpful for insightful and interpretable modeling and visualization of high-dimensional datasets. Despite having different dimension reduction methods in the literature, it is difficult to select the effective methods for the same, which necessitates quantifying the quality of the results they produce. Several metrics already exist in literature as well, but they are sometimes not applicable due to the huge space and time complexity. As a way out, an easy-to-handle metric is presented here, which can overcome the drawbacks of state-of-the-art metrics. In this study, a metric called CMET has been developed for quantifying the quality of a lower-dimensional embedding in terms of local ($CMET_L$) and global ($CMET_G$) shape-preserving capabilities of dimensionality reduction methods. Since both $CMET_L$ and $CMET_G$ account for the shape preservation capability of the underlying transformation, it is advisable to use the pair of scores ($CMET_L$, $CMET_G$) for deciding the actual degree of similarity in local and global shape that is retained in the embedding. The effectiveness of CMET has been demonstrated using three two-dimensional synthetic, one three-dimensional synthetic, two biological, and two image datasets, and fourteen dimensionality reduction methods have been used for this purpose.

At first, the embeddings of synthetic datasets have been visually examined with the original dataset, and then validated using CMET. For all three synthetic datasets, it has been observed, both visually and using CMET scores, that methods like PCA, SVD, and NMF, which project the data into a new space, performed well over methods like TSNE, ISOMAP, and UMAP, which try to recreate the data starting from random projection. For the three-dimensional synthetic dataset, PCA, ICA, and FA preserved both local and global shapes, and on the other hand, KPCA ruined both the local and global structures. Interestingly, both these facts have been supported by CMET scores. For the two biological datasets, nonlinear methods like UMAP, FITSNE, IVIS, and TSNE outperformed linear methods in terms of global shape preservation capability, although all these methods preserve local shapes quite well. These biological datasets have a very small number of samples than the number of features, which makes the task of dimensionality reduction even more difficult, but CMET has captured the shape preservation capability of all the methods efficiently. Similar results have been observed using state-of-the-art methods. Finally, two image datasets have been considered with even more samples. State-of-the-art methods fall short of being applied to these datasets due to their huge computational cost; however, CMET produces efficient scores. It has been observed, in most of the scenarios, that methods like LLE and LDA could not prove themselves to be very emulating competitors of the previously mentioned methods. It is also noteworthy to highlight that in both supervised and unsupervised fashion, CMET has yielded almost similar results. 

Later, a sensitivity analysis was performed for the parameter of CMET, and it was found to be not at all sensitive to hyper-parameters across all the datasets except the image datasets. More precisely, CMET has shown superior performance over the state-of-the-art metrics. For synthetic, biological, and image datasets, CMET scores support the visual representations as well. Thus, it may be mentioned here that CMET is a robust metric with less space and time complexity, and paves a smooth way for judging the performance of the dimension reduction methods.  


	
\section*{Authors' contributions statement}
Conceptualization of methodology and framework: SG, CM, RKD. Data curation, pre-processing, Implementation, and Initial draft preparation: SG, CM. Reviewing, Editing, and Overall Supervision: RKD.


\bibliography{mybib.bib}

\end{document}